\documentclass[journal]{IEEEtran}
\usepackage{array}
\usepackage{booktabs}
\usepackage{graphicx} 
\usepackage{subfigure} 
\usepackage{graphicx} 
\usepackage{amsfonts,amssymb}
\usepackage{multirow}
\usepackage{array}
\usepackage{booktabs} 
\usepackage{multirow}
\usepackage{url}

\usepackage{enumerate}
\usepackage{algorithm}
\usepackage{algorithmic}
\usepackage{amsmath}

\ifCLASSINFOpdf
\else
\fi
\begin{document}
%
\title{TEAM: We Need More Powerful Adversarial Examples for DNNs}
%
%
%

\author{Yaguan Qian,
        Ximin Zhang,
        Bin Wang$^{*}$,
        Wei Li,
       
        Zhaoquan Gu,
        Haijiang Wang,
        and Wassim Swaileh
       
\thanks{Yaguan Qian, Ximin Zhang and Wei Li are with the School of Big Data Science, Zhejiang University of Science and Technology, Hangzhou 310023, China (Email: qianyaguan@zust.edu.cn; 1289486793@qq.cn; liweikeyan@163.com). Bin Wang is with the Network and Information Security Laboratory of Hangzhou Hikvision Digital Technology Co., Ltd, Hangzhou 310052, China (Email: wbin2006@gmail.com). Zhaoquan Gu is with the Cyberspace Institute of Advanced Technology (CIAT), Guangzhou University, Guangzhou 510006, China (Email: zqgu@gzhu.edu.cn). Haijiang Wang is with the School of Information and Electronic Engineering, Zhejiang University of Science and Technology, Hangzhou 310023, China (Email: wanghaijiangyes@163.com). Wassim Swaileh is with the CY Cergy Paris University, ETIS Research Laboratory, Paris 95032, France (Email: wassim.swaileh@cyu.fr).}
}


%
%

\markboth{
}%
{Shell \MakeLowercase{\textit{et al.}}: Bare Demo of IEEEtran.cls for IEEE Journals}
%



\maketitle

\begin{abstract}

Although deep neural networks (DNNs) have achieved success in many application fields, it is still vulnerable to imperceptible adversarial examples that can lead to misclassification of DNNs easily. To overcome this challenge, many defensive methods are proposed. Indeed, a powerful adversarial example is a key benchmark to measure these defensive mechanisms. In this paper, we propose a novel method (TEAM, Taylor Expansion-Based Adversarial Methods) to generate more powerful adversarial examples than previous methods. The main idea is to craft adversarial examples by minimizing the confidence of the ground-truth class under untargeted attacks or maximizing the confidence of the target class under targeted attacks. Specifically, we define the new objective functions that approximate DNNs by using the second-order Taylor expansion within a tiny neighborhood of the input. Then the Lagrangian multiplier method is used to obtain the optimize perturbations for these objective functions. To decrease the amount of computation, we further introduce the Gauss-Newton (GN) method to speed it up. Finally, the experimental result shows that our method can reliably produce adversarial examples with 100\% attack success rate (ASR) while only by smaller perturbations. In addition, the adversarial example generated with our method can defeat defensive distillation based on gradient masking.

\end{abstract}


\IEEEpeerreviewmaketitle

\section{Introduction}

\IEEEPARstart{W}{ith} the rise of deep learning technology, DNNs have been successfully applied to many fields such as bioinformatics \cite{Chicco2,Spencer03}, speech recognition \cite{Mikolov04,Hinton05}, and computer vision \cite{LeCun06, LeCun07}. However, recent work shows that DNNs are vulnerable to \textit{adversarial examples}, which can fool DNNs by adding small perturbations to the input. Szeigy et al. \cite{SZEGEDY08} first reported the existence of adversarial examples in the image classification. Although these perturbations are almost imperceptible to human eyes, they can lead to disastrous consequences such as a traffic accident caused by an adversarial road sign image misclassified by autonomous driving systems. 

Adversarial attacks are divided into two categories, including untargeted attacks and targeted attacks. For untargeted attacks, the DNNs misclassify the adversarial examples to any class except for the ground-truth class, which include FGSM \cite{Lee09}, JSMA \cite{Nicolas10}, C\&W \cite{Carlini11}. For targeted attacks, the DNNs misclassify the adversarial examples to a specific target class, which include Deepfool \cite{Moosavi12} and M-DI$^{2}$-FGSM \cite{Xie14}. At the same time, many defensive methods are proposed as well, such as defensive distillation \cite{Papernot15}, adversarial training \cite{Madry13} , fortified networks \cite{Lamb16}, detecting adversarial examples \cite{Dathathri17} etc.

Up to date, adversarial training is shown as the most effective defensive method in practical experience \cite{Miyato18}. Aleksander et. al \cite{Madry13} consider adversarial training as a saddle point problem that is composed of an inner maximization and an outer minimization. The inner maximization aims to ﬁnd adversarial examples that cause the maximal loss of DNNs. Therefore, generating more powerful adversarial examples is a key component for effective adversarial training. At the same time, it can also be a key benchmark to measure current defensive methods.

Generating adversarial examples is usually modeled as a constrained optimization problem, which in general aims to maximize the loss function of DNNs. Different from previous work, we propose a novel approach, namely TEAM, to generate more powerful adversarial examples through minimizing the confidences of the ground-truth class in untargeted attacks or maximizing the confidences of the target class in targeted attacks, where confidences is the output of DNNs. Since DNNs are highly nonlinear, it easily falls into local minimum when using a confidence as an objective function. To obtain a smoother objective function, we use the second-order Taylor expansion to approximate the DNNs in a tiny neighborhood with regard to the input.  A typical gradient decrease method can be used to solve the above problem. As we know, unlike FGSM to use the first-order gradient, our method is the first attempt to generate adversarial examples using the second-order gradient.

\begin{figure}[t]

\centering

\subfigure[]{

\label{aa}
\includegraphics[width=8.5cm,height=6cm]{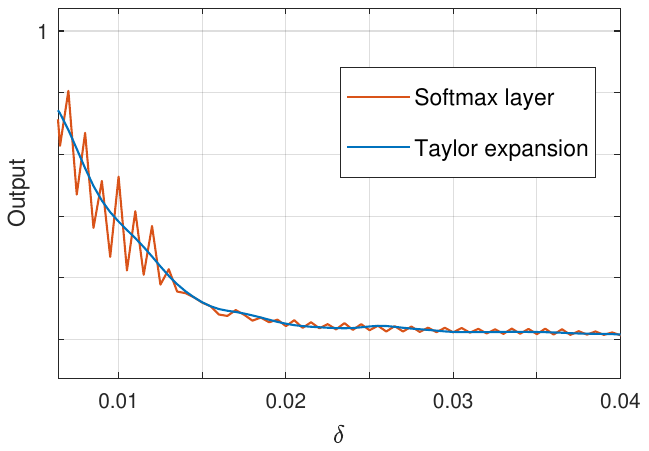}}

\hspace{1in}

\subfigure[]{
\label{bb}

\includegraphics[width=8.5cm,height=6cm]{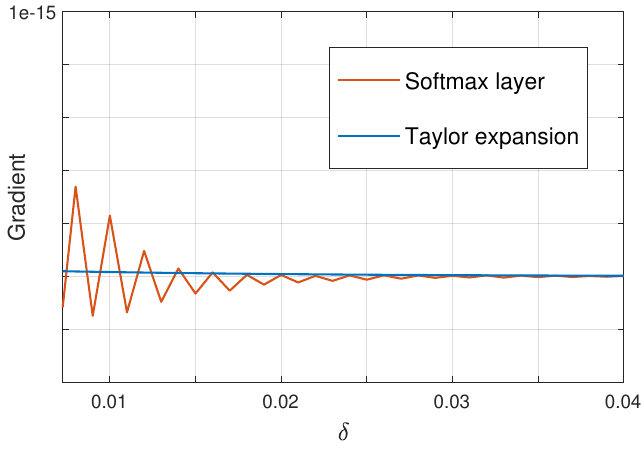}}

\caption{(a)the output of softmax layer to a specific class and that of the Taylor expansion to a specific class
     (b)the gradient of softmax layer to a specific class and that of the second-order Taylor expansion to a specific class}

\label{shattered}
\end{figure}

All current attack methods including ours rely on the gradient of DNNs, however,  previous attacks are based on the assumption that DNNs are smooth or convex \cite{Balduzzi19}. Unfortunately, DNNs are in general neither smooth nor convex and researchers may have underestimated the impact of this feature. In fact, the gradient of DNNs is extremely unsmooth as shown in Fig.\ref{shattered}. In Fig.\ref{aa}, the red line plots the confidence of a specific class calculated by the DNN and the blue line plots the confidence of the same class calculated by the second-order Taylor expansion of the DNN. Their corresponding gradients are shown in Fig.\ref{bb}. From Fig.\ref{shattered}, we can achieve two observations. The first is that the confidence of second-order Taylor expansion can perfectly approximate that of the DNN. The second is that the gradient of DNN is not easy for optimisation, which can easily lead to a local optimal solution \cite{Balduzzi19} and after using the second-order Taylor expansion to approximate it, it becomes smoother.

 Our work shows that using the second-order Taylor expansion to approximate the output of DNNs in a tiny neighborhood  can overcome the disadvantage of high nonlinearity of DNNs. After that, we can construct a dual problem with the Lagrangian multiplier method, which is a convex problem and we can hope to get the global optimum.It means that a more powerful adversarial example can be obtained with our second-order method than previous first-order method.

As we know, the calculation of a second-order matrix, known as the Hessian matrix, is usually with high computation costs. It is also not advisable to use a secant line approximation of the Hessian matrix. Therefore, in order to overcome the disadvantage of the Hessian matrix, we construct a least-squares cost function based on the DNN and use a first-order gradient matrix to substitute the Hessian matrix with the Gauss-Newton method.

The main contributions of this work are summarized as follows:
\begin{itemize}

\item Different from previous first-order methods, we propose a novel second-order method to craft adversarial examples by minimizing the confidence of the ground-truth class under untargeted attacks or maximizing the confidence of the target class under targeted attacks. Specifically, we define new objective functions that approximate DNNs by using the second-order Taylor expansion within a tiny neighborhood of the input, which can be easily optimized.
\item Our method can generate more powerful adversarial examples than previous work. The significance of our work lies in two aspects. The first is that more powerful adversarial examples contribute to more robust DNNs through adversarial training. The second is that a powerful adversarial example can be a significant benchmark to measure the state-of-the-art defensive mechanisms.
\item Compare to the state-of-the-art method C\&W, our generated adversarial example keep  less  likely  to  be  detected  by  human eyes with a higher ASR. On MNIST, ASR of our method achieves 3.8\% higher with $\ell_{2}$ = 1.84, PSNR = 67.21 than that of C\&W with $\ell_{2}$ = 2.21 and PSNR = 60.68. On CIFAR10, ASR of our method achieves 0.8\% higher with $\ell_{2}$ = 0.70 and PSNR = 72.86 than that of C\&W with $\ell_{2}$ = 0.75 and PSNR = 68.52 on averagely.
\end{itemize}

The rest of the paper is structured as follows: Sec.\ref{two} briefly describes the related work of this paper. Sec. \ref{three} introduces background on the fundamentals of DNNs and the most representative attack methods available nowadays. Sec. \ref{four} describes our method TEAM with detailed mathematical derivation. Sec.\ref{five} presents the experiment evaluation with two real datasets MNIST and CIFAR 10. In Sec. \ref{six} we conclude our work.

We share the source code of our approach online (https://github.com/zxm2020/the-Second-Order-Gradients-Based-Method-for-Generating-Adversarial-Examples).

\begin{figure}[t]

\begin{center}
\centerline{\includegraphics[width=8.5cm,height=8.5cm]{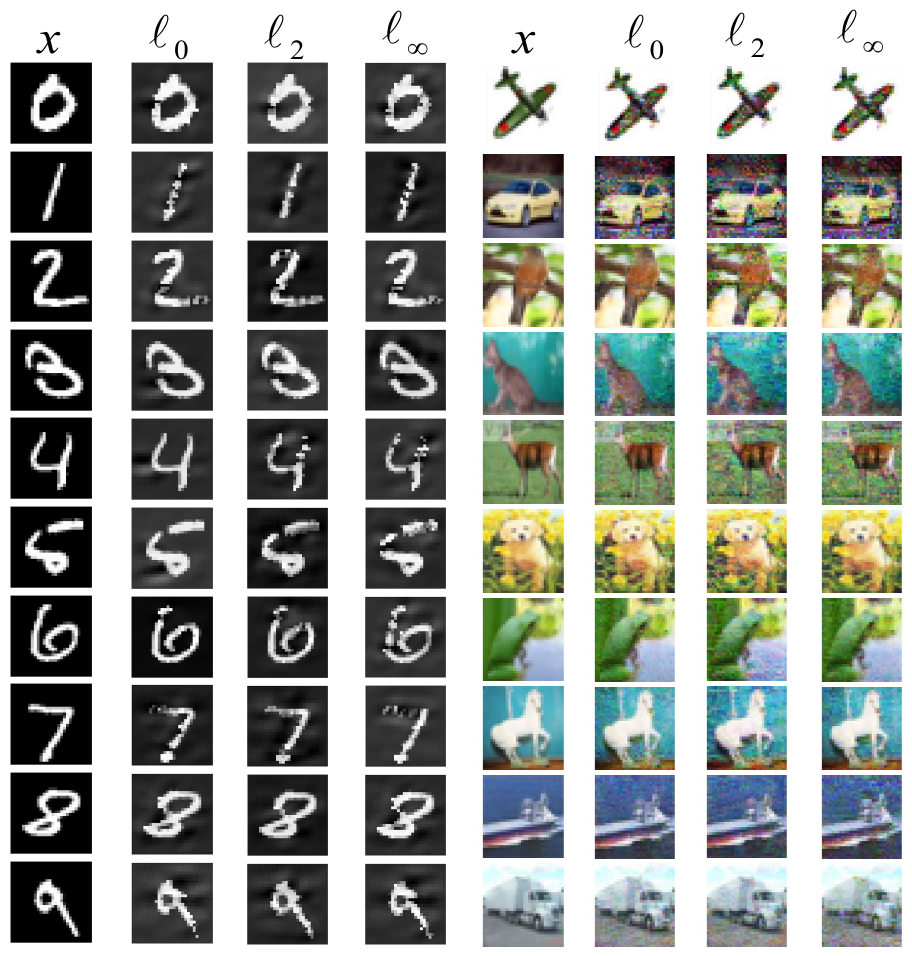}}
\caption{An illustration of our attacks on standard convolutional neural networks. The leftmost column is the original images. The next three columns show adversarial examples generated by our method with $\ell_{2}$, $\ell_{\infty}$ and $\ell_{0}$ constrains, respectively. All the original images are classified correctly, while the adversarial examples are misclassified to other classes. The images were chosen randomly from the test set of MNIST and CIFAR 10.}
\end{center}

\end{figure}


\section{RELATED WORK}
\label{two}

Since Szeigy et al. \cite{SZEGEDY08} discovered that DNN are vulnerable to adversarial examples, researchers proposed a variety of methods for generating adversarial examples, which can be summarized as the process of defining an objective function and then solving the optimization problem. 

Szegedy et al. \cite{SZEGEDY08} first proposed a box-constrained optimal perturbation method called L-BFGS. L-BFGS defined the mapping function of image to label as an objective function and used the linear search method to find adversarial examples based on it. Although L-BFGS is effective, the linear search method results in high computational complexity. Goodfellow et al. \cite{Lee09} proposed FGSM (Fast Gradient Sign Method) to reduce the computational complexity. FGSM constructs the objective function based on the gradient changes of loss function, which could perform one-step gradient update along the direction of the sign of gradient at each pixel, therefore the computation cost is extremely lower. However, FGSM is known as a weak attack method since one-step attack method like FGSM is easy to defend \cite{Kurakin21}. Based on FGSM, Kurakin et al. \cite{Bengio22} proposed BIM (Basic Iterstive Method) that makes multi-step small perturbations along the direction of gradient increase through iteration. Compared with FGSM, BIM could construct more accurate perturbations, but at the cost of larger computation. Further more, to improve BIM, Madry et al. \cite{Madry13} proposed PGD (Projected Gradient Descent) that takes the random perturbations of the original sample in its neighborhood as the initial input, and generates the adversarial examples after several iterations. In summary, all these methods defined a single objective function related to the loss function of DNNs and use different optimization methods based on the first-order gradient to calculate the optimum of the objective function.

Since then, researchers have proposed a variety of other improved algorithms. Papernot et al. \cite{Nicolas10} proposed JSMA (Jacobian-based Saliency Map Attack) for targeted attacks. The essence of this method is to iteratively modify the two most effective pixels to perturb based on the adversarial saliency map, repeating the process until misclassification is achieved. Besides, each iteration of JSMA needs to start a new recalculation of the gradient to generate a new adversarial saliency map. Compare to other methods, JSMA perturbs fewer pixels. Moosav-dezfooli et al. \cite{Moosavi12} proposed a new iterative attack method, Deepfool, which has similar effect of FGSM and can calculate smaller perturbations. By linearizing the boundary of the region where the image is located, Deepfool move the image gradually to the decision boundary, until the image is finally moved to the other side of the decision boundary, causing the classifier to classify incorrectly. Carlini and Wanger \cite{Carlini11} proposed the C\&W method, which defined a new objective function to optimize distance and the penalty term effectively. In addition, the authors introduced a new variant to avoid the box constraint, which enhances the accuracy of searching optimum but with a huge search cost. According to their further study \cite{Wagner23,Wagner24}, C\&W is effective for most of the existing adversarial detecting defenses. Xie et al. \cite{Xie14} applied momentum and diverse inputs to FGSM, which aims to improve transferability, namely M-DI$^{2}$-FGSM (Momentum Diverse Inputs Iterative Fast Gradient Sign Method).  In general, all these methods using the output vector of the DNN’s layer to construct objective functions, however, due to the non-convex nature of  DNNs, this methods are still easy to fall into local optimum.

However, due to the high nonlinearity of DNNs and the non-convex nature of their objective functions, these methods easily trap into local optimum. Although, through experiments, previous work has found that iterative methods like PGD could find effective local optimums \cite{Madry13}, and David Balduzzi et al. \cite{Balduzzi19} recognized the mathematical proof of the bounds relative to the global optimum as an impossibility, our method aims to find a fine-tuned local optimum theoretically and experimentally.

\section{BRACKGROUND}
\label{three}
This section introduces the primary notion relevant to our work,  including DNNs and classic attack algorithms.

\subsection{Deep Neural Networks and Notation}
\label{Notation}
A DNN is a multidimensional function $F(x): \mathbb{R}^{n} \mapsto \mathbb{R}^{m}$ that accepts an input  $x \in \mathbb{R}^{n}$  and produces an output $S \in \mathbb{R}^{m}$. $S$ is the output of the softmax layer and $ s_{i}$ is the $i$-element of $S$ , which satisfies $0 \leq s_{i} \leq 1$ and $s_{1}+\cdots+s_{m}=1$.  The DNN assigns the label $Y(x)=\arg \max _{i} F(x)_{i}$.  Let  $Y(x)=y$ be the ground-truth lable of $x$. The inputs of the softmax layer are called \textit{logits}.

An $N$-layer DNN accepts input $x$ and produces the  output $S$ as follows:
\begin{equation}S=\operatorname{softmax}\left(F^{(N)}\left(F^{(N-1)\ldots }\left(F^{(1)}(x)\right)\right)\right)
\label{DNN}
\end{equation}
where  $F^{(i+1)}(x)=\sigma\left(w_{i} \cdot F^{(i)}(x)\right)+b_{i}$ for non-linear activation function $\sigma(\cdot)$, model weights $w_{i}$ and model biases $b_{i}$. In our paper, $Z$ is the output of $F^{(N)}\left(F^{(N-1)\ldots }\left(F^{(1)}(x)\right)\right)$ (so $Z$ are  \textit{logits}) and $S=\operatorname{softmax}\left(Z\right)$.

\subsection{Adversarial Examples }
Szegedy et al. \cite{SZEGEDY08} first observed the existence of \textit{adversarial examples} that can be constructed from a valid input $x$ by solving the following optimization problem:

\begin{equation}
\label{eq2}
\operatorname{minimize} \mathcal{D}(x,x^{\prime})\  \  \text { s.t. }Y(x^{\prime})\neq y,  x^{\prime} \in[0,1]^{n}\end{equation}
where $x^{\prime}=x+\delta$ is the adversarial exsample and $ \mathcal{D}(\cdot)$ is a distance norm, which can be $\ell_{0}$,  $\ell_{2}$ or  $\ell_{\infty}$. 

Su et al.\cite{spot} established adversarial threat models in two dimensions, including adversarial goal and adversarial capabilities:

\textbf{Adversarial goal} is to construct an adversarial example $x^{\prime}$ that results in $Y(x^{\prime})\neq y$, which can be further divided into two categories:  (a) \textit{Untargeted attacks}: the adversary seeks to provide an input  $x^{\prime}$ that alter the output to any class different from $y$ , and (b) \textit{Targeted attacks}: the adversary seeks to provide an input $x^{\prime}$ that force the output to a specific target class $Y(x^{\prime})=y_{t}$.  In our work, both untargeted attacks and targeted attacks are discussed.

\textbf{Adversarial capabilities} represent how much information the adversary can obtain from the target DNN. In general, adversarial capabilities can be classified as white-box attacks and black-box attacks. \textit{White-box attacks} mean that the adversary knows almost everything about DNNs, including training data, activation functions, topology structures, weight coefficients and so on. \textit{Black-box attacks} assume that the adversary cannot obtain the internal information of the target DNN, except for $Z$ and $S$. In our work, we focus on white-box attacks.  

\subsection{Attack algorithms}
The existing typical gradient-based methods include FGSM, JSMA, Deepfool, C\&W, M-DI$^{2}$-FGSM, etc. which are used to compare with our method.

\textbf{FGSM}. Goodfellow et al. \cite{Lee09} presented  FGSM(Fast Gradient Sign Method) that performes one step gradient update along the direction of the sign of gradient at each pixel by restricting the  $\ell_{\infty}$-norm of the perturbations:

\begin{equation}
x^{\prime}=x+\epsilon \cdot \operatorname{sign}\left(\nabla_{x} J(F(x ; \theta), Y(x))\right)
\end{equation}
where $ J(\cdot)$ is the loss function, $\varepsilon$ is the magnitude of the perturbation. FGSM  has the advantages of low computational complexity and the ability to generate a large number of adversarial examples in a short time.

\textbf{JSMA}. Papernotet al. \cite{Nicolas10} proposed a targeted $\ell_{0}$  attack method by computing a saliency map using the gradients of the last hidden layer with respect to the input. For the target class $y_{t}$, adversarial saliency map $S(x,y_{t})$ is defined as :
\begin{equation}
\label{jsma}
S(x, y_{t})[i]=\left\{\begin{array}{l}
{0, \text { if } \frac{\partial Z_{y_{t}}}{\partial x}<0  \ \ o r \sum_{j \neq y_{t}} \frac{\partial Z_{j}}{\partial x}>0} \\
{\left(\frac{\partial Z_{y_{t}}}{\partial x}\right)\left|\sum_{j \neq y_{t}} \frac{\partial Z_{j}}{\partial x}\right|, \quad \text { otherwise }}
\end{array}\right.
\end{equation}In $S(x,y_{t})$, a larger value indicates a higher likelihood of fooling DNN. JSMA chooses the pixels that are most effective to fool the network and perturbs them.

\textbf{Deepfool}. Moosav-dezfooli et al. \cite{Moosavi12} proposed the Deepfool, which employs linear approximation for gradient iterative attack. At each iteration, Deepfool computes the minimal perturbations of  the linearlized DNN around the input as:

\begin{equation}\underset{\delta}{\arg \min }\left\|\delta\right\|_{2} \ \ \text { s.t. } F\left(x\right)+\nabla F\left(x\right)^{T} \delta=0\end{equation}The perturbations are accumulated to compute the final perturbation once $x^{\prime}$ is misclassified by DNN.

\textbf{C\&W}. Carlini and Wagner proposed a targeted  attack C\&W. Based on their further studies \cite{Carlini11,Wagner23,Wagner25}, C\&W  are effective against most existing defenses. They modeled the process of generating adversarial examples as the following optimization problem:
\begin{equation}\begin{array}{ll}
\operatorname{minimize} & d(x, x+\delta)+c \cdot f(x+\delta) \\
\text { s.t. } & x+\delta \in[0,1]^{n}
\end{array}\end{equation}
where $
f\left(x+\delta\right)=\max \left(\max _{i \neq y_{t}}\left(Z_{i}\right)-Z_{y_{t}},0\right)
$, and $d$ is the distance metric.

\textbf{M-DI$^{2}$-FGSM}. Xie et al. \cite{Xie14} combined momentum and diverse inputs naturally to form a much stronger attack,  Momentum Diverse Inputs Iterative Fast Gradient Sign Method (M-DI$^{2}$-FGSM). By performing random image re-sizing and padding as image transformation$T(\cdot)$:
\begin{equation}g_{n+1}=\mu \cdot g_{n}+\frac{\nabla_{x} L\left(T\left(x_{n} ; p\right), y ; \theta\right)}{\left\|\nabla_{x} L\left(T\left(x_{n} ; p\right), y ; \theta\right)\right\|_{1}}\end{equation}
\begin{equation}
x^{n+1}=\operatorname{Clip}_{x}^{\varepsilon}\left\{x^{n}+\alpha \cdot \operatorname{sign}\left(g_{n+1})\right)\right\}
\end{equation}where $\mu$ is the decay factor of the momentum term and $g_{n}$ is the accumulated gradient.

\section{OUR APPROACH}
\label{four}
The problem of constructing adversarial examples can be formally defined as problem (\ref{eq2}), yet it is is difficult for existing algorithms to solve directly, as the properties of DNN make it highly non-linear and non-convex. Therefore, we express problem (\ref{eq2}) in a different form, which is better adapted for optimization. In addition, the Lagrangian multiplier method is introduced to solve this problem. 

\subsection{Problem Formulation}
Remind that $Z$ are logits, $Z_{i}$ is the logits element of class $i$, $S=\operatorname{softmax}\left(Z\right)$, and  $S_{i}$ is the confidence of class $i$. A larger value of $Z_{i}$ or $S_{i}$ indicates a higher likelihood that the DNN classify the input as class $i$ and vice versa. In our work, we aim to minimize $Z_{y}$ or $S_{y}$ in untargeted attacks, which is equivalent to minimizing the likelihood of proper classification. In targeted attacks, we aim to maximize $Z_{y_{t}}$ or $S_{y_{t}}$, which is equivalent to maximizing the likelihood of forcing the DNN to classify the input as  target class $y_{t}$. 

Although $Z_{i}$ and $S_{i}$ are seemed equally, there are some differences between $Z_{i}$ and $S_{i}$. Since $S_{i}=\operatorname{softmax}\left(Z_{i}\right)=\exp \left(Z_{i}\right) / \sum_{j=1}^{m} \exp \left(Z_{j}\right)$,  $S_{i}$ satisfies $0 \leq S_{i} \leq 1$ and $S_{1}+\cdots+S_{m}=1$, which indicates that the decrease of $S_{y}$ corresponds to the increase of $\sum_{i \neq y}S_{i}$ and vice versa. Therefore,  the decrease of $S_{y}$ ensures higher likelihood of misclassification and the increase of $S_{y_{t}}$  ensures higher likelihood of forcing the DNN to classify the input as  class $y_{t}$. But Papernotet al. \cite{Nicolas10}  claimed that $S_{i}$ will introduce extreme derivative values through $S_{i}=\operatorname{softmax}\left(Z_{i}\right)$. Compared with $S$,  the decrease of  $Z_{y}$ cannot guarantee the higher value of $\sum_{i \neq y} Z_{i}$,  it is unaffected by extreme derivative values.  In order to get deep insight to the differences between $z_{i}$ and $s_{i}$, we construct adversarial examples based on $Z_{i}$ and $S_{i}$ respectively.

Due to DNNs' highly non-linear and non-convex, to solve the problem (2) is non-trivial \cite{jsma25}. Hence, we define some objective functions $T_{i}$, $i$=1,2,...,6 suitable for optimization. In untargeted attacks,  we minimize $T_{i}$ such that the likelihood of $Y(x^{\prime})\neq y$ will be maximized. In targeted attacks, we maximize $T_{i}$ such that the likelihood of $Y(x^{\prime})=y_{t}$ will be maximized. The corresponding $T_{i}$ is defined as following:\\

      \quad \quad \quad \quad \quad \quad  $T_{1}(x^{\prime})=Z_{y}(x^{\prime})$ \\

      \quad \quad \quad \quad \quad \quad  $T_{2}(x^{\prime})=Z_{y_{t}}(x^{\prime})$ \\
      
      \quad \quad \quad \quad \quad \quad    $T_{3}(x^{\prime})=Z_{y_{t}}(x^{\prime})-\max _{i \neq y_{t}}\left(Z_{i}(x^{\prime})\right)$\\
      
      \quad \quad \quad \quad \quad \quad    $T_{4}(x^{\prime})=S_{y}(x^{\prime})$ \\
      
      \quad \quad \quad \quad \quad \quad    $T_{5}(x^{\prime})=S_{y_{t}}(x^{\prime})$\\
      
      \quad \quad \quad \quad \quad \quad    $T_{6}(x^{\prime})=S_{y_{t}}(x^{\prime})-\max _{i \neq y_{t}}\left(S_{i}(x^{\prime})\right)$

We use the alternative formulation for untargeted attacks:

\begin{equation}
\label{objective1}
\begin{aligned}
\operatorname{minimize} &\ \ T_{1}(x+\delta)\  \text { or }\  T_{4}(x+\delta) \\
\ \ \text { s.t. } & \ F(x+\delta) \neq y \\
&\  \ x+\delta \in[0,1]^{n}\\
&\  \ \mathcal{D}(x,x+\delta)<C
\end{aligned}\end{equation}
and the alternative formulation for targeted attacks:

\begin{equation}
\label{objective2}
\begin{array}{rl}
\operatorname{maximize} & T_{2}(x+\delta) \text { or } T_{3}(x+\delta)  \text { or } T_{5}(x+\delta) \text { or } T_{6}(x+\delta) \\
\ \text { s.t. } &F(x+\delta)=y_{t} \\
& x+\delta \in[0,1]^{n}\\
& \mathcal{D}(x,x+\delta)<C
\end{array}\end{equation}where $C$ is a perturbation bound of $\delta$, $\mathcal{D}$ is either $\ell_{0}$, $\ell_{2}$ or $\ell_{\infty}$, i.e., $\mathcal{D}(x,x+\delta)=\|\delta\|_{p}$.

\subsection{Generate adversarial examples based on the Second-order Taylor expansion}
\label{4.2}
In this section, we address how TEAM solve problem (\ref{objective1}) and  problem (\ref{objective2}).  First, we adopt the quadratic approximation to the nonlinear part of $T_{i}$ in a tiny neighborhood of $x$  by the second-order Taylor Expansion. At this point, the original non-trivial nonlinear optimization problem can be transformed into a second-order constrained optimization problem. To solve this, we use the Lagrangian multiplier method to construct a dual problem, which is convex and provides an approximately optimal solution to the original  problem (\ref{objective1}) and (\ref{objective2}). 

For the untargeted attack problem (\ref{objective1}),  we first compute the gradient matrix $\nabla_{x} S_{y}(x)$, $\nabla_{x} Z_{y}(x)$and the Hessian matrix $\nabla_{x}^{2} S_{y}(x)$, $\nabla_{x}^{2} Z_{y}(x)$ for the given example $x$.\begin{equation}\nabla_{x} S_{y}(x)=\left[\frac{\partial Z_{y}(x)}{\partial x_{i}}\right]_{n \times 1}\end{equation}
\begin{equation}\nabla_{x} Z_{y}(x)=\left[\frac{\partial Z_{y}(x)}{\partial x_{i}}\right]_{n \times 1}\end{equation}
\begin{equation}\nabla_{x}^{2} S_{y}(x)=\left[\frac{\partial^{2} S_{y}(x)}{\partial x_{i} \partial x_{j}}\right]_{n \times n}\end{equation}
\begin{equation}\nabla_{x}^{2} Z_{y}(x)=\left[\frac{\partial^{2} Z_{y}(x)}{\partial x_{i} \partial x_{j}}\right]_{n \times n}\end{equation}We calculate the partial derivatives of the DNN directly, rather than its cost function, and we differentiate with respect to $x$ rather than the parameters of the DNN. As a result, this allows us to approximate $S_{y}(x)$ and $Z_{y}(x)$ in a tiny neighborhood of $x$. Our goal is to express $\nabla_{x} S_{y}(x)$, $\nabla_{x} Z_{y}(x)$, $\nabla_{x}^{2} S_{y}(x)$ and $\nabla_{x}^{2} Z_{y}(x)$ according to one input dimension $x_{i}$. We start at softmax layer :\begin{equation}\frac{\partial S_{y(x)}}{\partial x_{i}}=\sum_{j=1}^{m}\left[\frac{\partial S_{y}}{\partial Z_{j}} \cdot \frac{\partial Z_{j}}{\partial x_{i}}\right]
\label{gradient1}
\end{equation} Since $S_{y}=\operatorname{softmax}\left(Z_{y}\right)=\exp \left(Z_{y}\right) / \sum_{p=1}^{m} \exp \left(Z_{p}\right)$,  Eq.(\ref{gradient1}) yields:
\begin{equation}
\label{grad2}
\begin{aligned}
\frac{\partial S_{y(x)}}{\partial x_{i}} &=\sum_{j=1}^{m}\left[\frac{\partial \frac{\exp \left(Z_{y} \right)}{\sum_{p=1}^{m} \exp \left(Z_{p}\right)}}{\partial Z_{j}} \cdot \frac{\partial Z_{j}}{\partial x_{i}}\right] \\
&=\sum_{j \neq y}\left[\frac{-\exp \left(Z_{y}\right)   \exp \left(Z_{j}\right)}{\left(\sum_{p=1}^{m} \exp \left(Z_{p}\right)\right)^{2}} \cdot \frac{\partial Z_{j}}{\partial x_{i}}\right] \\
& +\frac{\exp \left(Z_{y}\right)   \sum_{p=1}^{m} \exp \left(Z_{p}\right)-\left(\exp \left(Z_{y}\right)\right)^{2}}{\left(\sum_{p=1}^{m} \exp \left(Z_{p}\right)\right)^{2}} \cdot \frac{\partial Z_{y}}{\partial x_{i}} \\
&=\sum_{j \neq y}\left(-S_{y}   S_{j}   \frac{\partial Z_{j}}{\partial x_{i}}\right)+\left[S_{y}-\left(s_{y}\right)^{2}\right]   \frac{\partial Z_{y}}{\partial x_{i}}
\end{aligned}
\end{equation}In Eq.(\ref{grad2}), according to Section\ref{Notation}, all terms are known except for $ \frac{\partial Z_{y}}{\partial x_{i}}$. By calculating $Z_{y}$, we have

\begin{equation}
\label{grad3}
\begin{aligned}
\frac{\partial Z_{y(x)}}{\partial x_{i}} &=\frac{\partial \sigma_{N,y}\left(w_{N,y} \cdot H^{N-1}+b_{N,y}\right)}{\partial x_{i}} \\
&=\frac{\partial \sigma_{N,y}\left(w_{N,y} \cdot H^{N-1}+b_{N,y}\right)}{\partial x_{i}}  \\
& \quad \cdot   w_{N, y}   \frac{\partial H^{N-1}}{\partial x_{i}}
\end{aligned}
\end{equation}where $H^{N-1}$ is the output vector of layer $N-1$ and $z=H^{N}$. Besides, $\sigma_{N, y}(\cdot)$ is the activation function of neuron $y$ in layer $N$.  $w_{N, y}$ and $b_{N, y}$ is the weights and bias for neuron $y$ in layer $N$. In this equation, $ \frac{\partial H^{N-1}}{\partial x_{i}} $ is the only unknown element that we are able to compute recursively. Therefore, we get $\nabla_{x} S_{y}(x)$ and $ \nabla_{x} Z_{y}(x)$ by plugging these results for hidden layers back into Eq.(\ref{grad2}) and Eq.(\ref{grad3}).

$\nabla_{x}^{2} S_{y}(x)$ and $\nabla_{x}^{2} Z_{y}(x)$ are calculated based on $\nabla_{x} S_{y}(x)$ and $ \nabla _{x}Z_{y}(x)$:
\begin{equation}
\label{grad4}
\begin{aligned}
\nabla_{x}^{2} S_{y}(x) &=\frac{\partial^{2} S_{y}(x)}{\partial x_{i} \partial x_{j}} \\
&=\frac{\partial}{\partial x_{j}}\left[\frac{\partial S_{y}(x)}{\partial x_{i}}\right] \\
&=\frac{\partial}{\partial x_{j}}\left[\sum_{j \neq y}\left(-S_{y} S_{j}   \frac{\partial Z_{j}}{\partial x_{i}}\right) \right. \\
&+ \left.\left(S_{y}-\left(S_{y}\right)^{2}\right)   \frac{\partial Z_{j}}{\partial x_{i}}\right] \\
&=\sum_{j \neq y}\left(-S_{y} S_{j}   \frac{\partial^{2} Z_{j}}{\partial x_{i} \partial x_{j}}\right) \\
& +  \left(S_{y}-\left(S_{y}\right)^{2}\right)   \frac{\partial^{2} Z_{j}}{\partial x_{i} \partial x_{j}}
\end{aligned}\end{equation}
\begin{equation}
\label{grad5}
\begin{aligned}
\nabla_{x}^{2} Z_{y}(x)=& \frac{\partial^{2} Z_{y}(x)}{\partial x_{i} \partial x_{j}} \\
=& \frac{\partial}{\partial x_{j}}\left[\frac{\partial Z_{y}(x)}{\partial x_{i}}\right] \\
=& \frac{\partial}{\partial x_{j}}\left[\frac{\partial \sigma_{N, y}\left(w_{N, y}  H^{N-1}+b_{N, y}\right)}{\partial x_{i}} \right.\\
&\cdot \left.w_{N, y}   \frac{\partial H^{N-1}}{\partial x_{i}}\right] \\
=&  \frac{\partial^{2} \sigma_{N, y}\left(w_{N, y}  H^{N-1}+b_{N, y}\right)}{\partial x_{i} \partial x_{j}} \\
&   \cdot w_{N, y}  \frac{\partial^{2} H^{N-1}}{\partial x_{i} \partial x_{j}}
\end{aligned}\end{equation}In Eq.(\ref{grad4}) and Eq.(\ref{grad5}), all terms are known except for $\frac{\partial^{2} Z_{j}}{\partial x_{i} \partial x_{j}}$ and $\frac{\partial^{2} H^{N-1}}{\partial x_{i} \partial x_{j}}$, which can be calculated recursively. By plugging these results into  Eq.(\ref{grad4}) and Eq.(\ref{grad5}), we obtain $\nabla_{x}^{2} S_{y}(x)$ and $\nabla_{x}^{2} Z_{y}(x)$.

Second, we use the second-order Taylor Expansion to approximate the value of $S_{y}(x)$ and $Z_{y}(x)$ in the neighborhood $U({x}, \delta)$. Since $S_{y}$ and $Z_{y}$ contribute equally and can be substituted for each other completely in the rest of the process, we use the same symbol $f_{y}$ to denote both $S_{y}$ and $Z_{y}$, which is easily understood.
\begin{equation}
\begin{aligned}
f_{y}\left(x^{\prime}\right) &=f_{y}(x+\delta) \\
& \approx  \phi_{y}(\delta) \\
&=f_{y}(x)+\nabla f_{y}(x)^{T}   \delta+\frac{1}{2}   \delta^{T}  \nabla^{2} f_{y}(x)  \delta
\end{aligned}
\end{equation}
$x^{\prime}$ is the moving point in the neighborhood $U(x, \delta)$ and $\phi_{y}(\delta)$ gives the second-order approximation of $f_{y}\left(x^{\prime}\right)$. 

Finally, for the untargeted attack problem (\ref{objective1}), we transform it into:
\begin{equation}
\label{problem1}
\textup{minimize }\phi_{y}(\delta) \quad \text { s.t. }\|\delta\|_{p} \leq C\end{equation}
We construct the Lagrangian function:

\begin{equation}
\label{problem2}
\begin{aligned}
P(\delta, \lambda) &=\phi_{y}(\delta) +\lambda\left(\|\delta\|_{p}-C\right) \\
&=f_{y}(x)+\nabla f_{y}(x)^{T}   \delta \\
&+\frac{1}{2} \delta^{T} \nabla^{2} f_{y}(x)   \delta \\
&+\lambda\left(\|\delta\|_{p}-C\right)
\end{aligned}\end{equation}Now, we  transformed the optimization problem (\ref{problem1}) with inequality constraints into an unconstrained optimization problem:
\begin{equation}
\label{origin}
\textup{minimize }\ P(\delta, \lambda)
\end{equation}To solve this optimization problem, we have the KKT condition:
\begin{equation}\left\{\begin{array}{l}
\nabla_{\delta} P(\delta, \lambda)=\overrightarrow{0} \\
\nabla_{\lambda} P(\delta, \lambda)=\overrightarrow{0} \\
\|\delta\|_{p} \leq C \\
\lambda \geq 0 \\
\lambda\left(\|\delta\|_{p}-C\right)=0
\end{array}\right.\end{equation}And the dual problem of the problem (\ref{origin}), which  is a convex optimization problem and providing an approximately optimal solution to the original problem (\ref{objective1}):
\begin{equation}\max _{\lambda} \min _{\delta} P( \delta,\lambda) \quad \text { s } t . \lambda \geq 0\end{equation}According to the principle of weak duality property:  (a) the optimal value of the primal problem is not less than the optimal value of the dual problem; (b) the primal problem is convex and satisfies the Slater condition, the optimal value of the dual problem equals to the optimal value of the primal problem. However, due to the nonlinearity of DNNs, $\nabla_{x}^{2}f_{y}(X)$ is difficult to be proved as a positive definite matrix, which means the problem (\ref{origin}) cannot be a convex problem. Therefore, we think of the optimal value$\left(\delta^{*}, \lambda^{*}\right)$ of the dual problem as an approximately optimal solution to the primal problem (\ref{origin}). Our method to generate adversarial examples is presented in Algorithm\ref{algorithm1}. 

\begin{figure}[t]

\begin{center}

\centerline{\includegraphics[width=8cm]{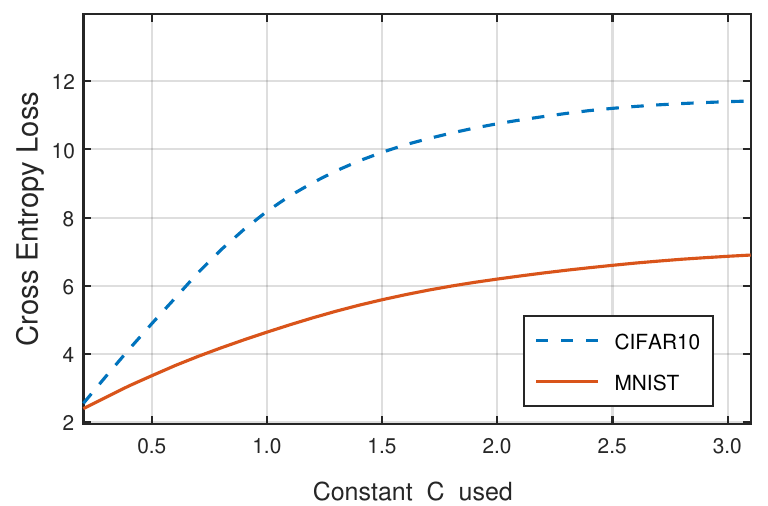}}
\caption{We present the relationship between the perturbation bound $C$ and the change of cross entropy loss of the DNN, where the adversarial examples are generated with the objective function $T_{1}$ on MNIST and CIFAR10 datasets respectively.}
\label{constant}
\end{center}
\end{figure}

\begin{algorithm}[t] 
\caption{ Generate adversarial examples based on the Taylor expansion for untargeted attack} 
\label{algorithm1} 
\begin{algorithmic}[1] 
\REQUIRE $x$, $y$, $C$,  

\ENSURE  $x^{\prime}$

\STATE 
$C\leftarrow 0.5$, $l\leftarrow y$ 
\STATE $\nabla _{x}f_{y}(x)\leftarrow\left[\frac{\partial f_{y}(x)}{\partial x_{i}}\right]_{n \times 1}$ \\
\STATE$ \nabla_{x}^{2} f_{y}(x) \leftarrow\left[\frac{\partial^{2} f_{y}(x)}{\partial x_{i} \partial x_{j}}\right]_{n\times n }$
\label{ code:fram:extract }

\STATE $\textbf {while}$\ $ l=y$  $\textbf {do}$
\label{code:fram:trainbase}
\STATE \ \ \ $\phi_{y}(\delta)=f_{y}(x)+\nabla_{x} f_{y}(x)^{T}  \delta+\frac{1}{2} \delta^{T}  \nabla_{x}^{2} f_{y}(x)  \delta$
  \ \ \      //Use \\  \ \ the Taylor expansion to approximate $f_{y}(x+\delta) $ in the  \\ 
  \ \ tiny neighbourhood of $x$
\STATE  \ \  $P(\delta, \lambda)=\phi_{y}(\delta)+\lambda \left(\|\delta\|_{p}-C\right)$
  \ \ //Construct the \\
  \ \ Lagrangian function\\
   
\STATE  \ \  $(\delta^{*},\lambda^{*})\leftarrow\max _{\lambda} \min _{\delta} P( \delta,\lambda) \quad \text { s.t.}  \lambda \geq 0$ //Calculate  \\
\ \ the dual problem\\

\STATE \ \ \  $l \leftarrow Y\left(X+\delta^{*}\right)$

\STATE \ \ \ $C \leftarrow C+0.1$
\STATE $\textbf {end while}$
\STATE $\textbf {return}$\ \ $x+\delta^{*}$

\end{algorithmic}
\end{algorithm}

In addition, the selection of perturbation bound $C$ is also involved in solving the problem (\ref{problem1}). The larger $C$ leads to the higher ASR, however the adversarial example will be perceptible easily. Therefore, the determination of $C$ is crucial to the effectiveness of adversarial examples. Empirically, the most suitable $C$ is the minimum one which satisfies $ f_{y_{t}}(x+\delta^{*})>f_{y}(x+\delta^{*})$ after solving problem (\ref{origin}). Fig.\ref{constant} shows relationship between $C$ and the change of cross entropy loss value where $C$ changed in the range [0, 3]. 


For targeted attacks(\ref{objective2}), we first compute the gradient matrix $\nabla_{x} f_{y_{t}}(x)$ and the Hessian matrix $\nabla_{x}^{2}f_{y_{t}}(x)$ for the target class $y_{t}$.\begin{equation}\nabla_{x} f_{y_{t}}(x)=\left[\frac{\partial f_{y_{t}}(x)}{\partial x_{i}}\right]_{n \times 1}\end{equation}
\begin{equation}\nabla_{x}^{2} f_{y_{t}}(x)=\left[\frac{\partial^{2} f_{y_{t}}(x)}{\partial x_{i} \partial x_{j}}\right]_{n \times n}\end{equation}
 $\nabla_{x} f_{y_{t}}(x)$ and $\nabla_{x}^{2}f_{y_{t}}(x)$ have the same derivation with  $\nabla _{x}f_{y}(x)$ and $\nabla_{x}^{2}f_{y}(x)$, which replace $y$ with $y_{t}$. Then, we use the second-sorder Taylor Expansion to approximate the value of $S_{y_{t}}$ and $Z_{y_{t}}$ in the tiny neighborhood of $x$:

\begin{equation}
\begin{aligned}
f_{y_{t}}\left(x^{\prime}\right) &=f_{y_{t}}(x+\delta) \\
& \approx  \phi_{y_{t}}(\delta) \\
&=f_{y_{t}}(x)+\nabla _{x}f_{y_{t}}(x)^{T}  \delta+\frac{1}{2}\delta^{T}  \nabla_{x}^{2} f_{y_{t}}(x) \delta
\end{aligned}
\end{equation}
where $\phi_{y_{t}}(\delta)$ gives the second order approximation of $f_{y_{t}}\left(x^{\prime}\right) $. Then, we transformed problem (\ref{objective2}) into:\begin{equation}
\label{T2}
\textup{maximize }\ \phi_{y_{t}}(\delta) \quad \text { s.t. }\|\delta\|_{p} \leq C\end{equation}and\begin{equation}
\label{T3}
\textup{maximize }\ \left[\phi_{y_{t}}(\delta)-\max _{i \neq y_{t}}\left(\phi_{i}(\delta)\right)\right] \quad \text { s.t. }\|\delta\|_{p} \leq C\end{equation}To solve the problem (\ref{T2}), we construct the Lagrangian function:
\begin{equation}
\label{Q1}
\begin{aligned}
Q_{1}(\delta, \lambda) &=\phi_{y_{t}}(\delta) +\lambda\left(\|\delta\|_{p}-C\right) \\
&=f_{y_{t}}(x)+\nabla_{x} f_{y_{t}}(x)^{T}   \delta \\
&+\frac{1}{2}   \delta^{T} \nabla_{x}^{2} f_{y_{t}}(x)   \delta \\
&+\lambda\left(\|\delta\|_{p}-C\right)
\end{aligned}\end{equation}Next, we transformed the problem (\ref{T2})  into an  unconstrained optimization problem by substituting Eq.(\ref{Q1}) in it:
\begin{equation}
\label{origin1}
\textup{maximize}\ Q_{1}(\delta, \lambda)\end{equation}The KKT condition of problem (\ref{origin1}) must be satisfied as follow:\begin{equation}\left\{\begin{array}{l}
\nabla_{\delta} Q_{1}(\delta, \lambda)=\overrightarrow{0} \\
\nabla_{\lambda} Q_{1}(\delta, \lambda)=\overrightarrow{0} \\
\|\delta\|_{p} \leq C \\
\lambda \geq 0 \\
\lambda\left(\|\delta\|_{p}-C\right)=0
\end{array}\right.\end{equation}And the dual problem of problem (\ref{origin1}):
\begin{equation}
\label{dual1}\min _{\lambda} \max _{\delta} Q_{1}( \delta,\lambda) \quad \text { s } t . \lambda \geq 0\end{equation}

According to the principle of weak duality property, the optimal solution obtained from the problem (\ref{dual1}) is the approximately optimal solution to the problem (\ref{origin1}). The algorithm is presented in Algorithm\ref{algorithm2}.

To solve the problem (\ref{T3}), we construct the Lagrangian function:
\begin{equation}
\label{Q2}
\begin{aligned}
Q_{2}(\delta, \lambda) &=\phi_{t}(\delta)-\max _{i \neq y_{t}}\left(\phi_{i}(\delta)\right) +\lambda\left(\|\delta\|_{p}-C\right) \\
&=f_{y_{t}}(x)+\nabla_{x} f_{y_{t}}(x)^{T}   \delta \\
&+\frac{1}{2}   \delta^{T}   \nabla_{x}^{2} f_{y_{t}}(x)   \delta \\
&+\ max_{i \neq y_{t}}[ \ f_{i}(x)+\nabla_{x} f_{i}(x)^{T}   \delta \\
&+\frac{1}{2}  \delta^{T}   \nabla_{x}^{2} f_{i}(x)   \delta \  ] \\
&+\lambda\left(\|\delta\|_{p}-C\right)
\end{aligned}\end{equation}
After substituting Eq.(\ref{Q2}) into problem (\ref{T3}), we have an unconstrained optimization problem:\begin{equation}
\label{origin2}
\textup{maximize}\ Q_{2}(\delta, \lambda)\end{equation} The KKT condition of problem (\ref{dual2}) must be satisfied as follow:\begin{equation}\left\{\begin{array}{l}
\nabla_{\delta} Q_{2}(\delta, \lambda)=\overrightarrow{0} \\
\nabla_{\lambda} Q_{2}(\delta, \lambda)=\overrightarrow{0} \\
\|\delta\|_{p} \leq C \\
\lambda \geq 0 \\
\lambda\left(\|\delta\|_{p}-C\right)=0
\end{array}\right.\end{equation}And dual problem of problem (\ref{origin2}):
\begin{equation}
\label{dual2}\min _{\lambda} \max _{\delta} Q_{2}( \delta,\lambda) \quad \text { s } t . \lambda \geq 0\end{equation}According to the principle of weak duality property, the optimal solution obtained from the problem (\ref{dual2}) is the approximately optimal solution to the problem (\ref{origin2}). Algorithm\ref{algorithm2} shows the process of targeted attacks.

\begin{algorithm}[t] 
\caption{ Generate adversarial examples based on the Taylor expansion for problem (\ref{T2}) and (\ref{T3})} 
\label{algorithm2} 
\begin{algorithmic}[1] 
\REQUIRE $x$, $y$, $C$ 

\ENSURE  $x^{\prime}$
\STATE $C\leftarrow 0.5$, $l\leftarrow y$ 
\STATE $\nabla f_{y_{t}}(x)\leftarrow\left[\frac{\partial f_{y_{t}}(x)}{\partial x_{i}}\right]_{n \times 1}$ 
\STATE $ \nabla^{2} f_{y_{t}}(x) \leftarrow\left[\frac{\partial^{2} f_{y_{t}}(x)}{\partial x_{i} \partial x_{j}}\right]_{n\times n }$

\STATE $\textbf {while}$\ $ l \neq y_{t}$ $\textbf {do}$

\STATE \ \ \ $\phi_{t}(\delta)=f_{y_{t}}(x)+\nabla_{x} f_{y_{t}}(x)^{T}   \delta+\frac{1}{2}   \delta^{T}   \nabla_{x}^{2} f_{y_{t}}(x)  \delta$  \       //Use \\
\ \ the Taylor expansion to approximate $f_{y_{t}}(x+\delta) $ in  the \\
\ \ tiny neighbourhood of $x$

\STATE \ \ $\left(\delta^{*}, \lambda^{*}\right) \leftarrow \label{dual1}\min _{\lambda} \max _{\delta} Q_{i}( \delta,\lambda)$, $i \in\{1,2\}$  $\text { s.t.}  \lambda \geq 0$  \\
\ \ \  //Calculate the dual problem\\

\STATE \ \ \ $l$ $\leftarrow Y\left(X+\delta^{*}\right)$

\STATE \ \ \ $C \leftarrow C+0.1$
\STATE $\textbf {end while}$
\STATE $\textbf {return}$\ \ $x+\delta^{*}$

\end{algorithmic}
\end{algorithm}

\subsection{Generate Adversarial Examples Based On The Gauss-Newton Method}
\label{4.3}
Since Sec.~\ref{4.2} involves the calculation of the Hessian matrix which requires a large amount of calculation, this section describe the GN method for improvement. The GN is a specialized method for minimizing the least-squares cost $(1 / 2)\left\|f_{y_{t}}(x+\delta)\right\|^{2}$. Given a point $x$ , the pure form of the GN method is based on linear function $f_{y_{t}}(x+\delta)$ to obtain

\begin{equation}
R\left(x+\delta, x+\delta^{k}\right)=f_{y_{t}}(x)+\nabla f_{y_{t}}(x)^{T}\left(\delta-\delta^{k}\right)
\end{equation}
and then minimizing the norm of the linear function $R$:
\begin{equation}
\begin{aligned}
\delta^{k+1}=& \arg \min _{\delta \in \mathbb{R}^{m}}\frac{1}{2}\left\|R\left(x+\delta, x+\delta^{k}\right)\right\|^{2} \\
=& \arg \min _{\delta \in \mathbb{R}^{m}}\frac{1}{2}\left\{2\left(\delta-\delta^{k}\right)^{T} \nabla f_{y_{t}}(x) f_{y_{t}}(x)\right.\\
&+\left(\delta-\delta^{k}\right)^{T} \nabla f_{y_{t}}(x) \nabla f_{y_{t}}(x)^{T}\left(\delta-\delta^{k}\right) \\
&\left.+\left\|f_{y_{t}}(x)\right\|^{2}\right\}
\end{aligned}
\end{equation}
Assuming that the matrix $\nabla f_{y_{t}}(x) \nabla f_{y_{t}}(x)^{T}$ is invertible, the above quadratic minimization yields:
\begin{equation}
\label{gaosi}
\delta^{k+1}=\delta^{k}-\left(\nabla f_{y_{t}}(x) \nabla f_{y_{t}}(x)^{T}\right)^{-1} \nabla f_{y_{t}}(x) f_{y_{t}}(x)
\end{equation}
Notice the high nonlinearity of DNNs, we cannot prove the matrix $\nabla f_{y_{t}}(x) \nabla f_{y_{t}}(x)^{T}$ is invertible. To ensure descent, and also to deal with the case where the matrix $\nabla f_{y_{t}}(x) \nabla f_{y_{t}}(x)^{T}$ is singular (as well as enhance convergence when this matrix is nearly singular), the Eq.(\ref{gaosi}) is rewrited as follows:
\begin{equation}
\delta^{k+1}=\delta^{k}-\alpha^{k}\left(\nabla f_{y_{t}}(x) \nabla f_{y_{t}}(x)^{T}+\Delta^{k}\right)^{-1} \nabla f_{y_{t}}(x) f_{y_{t}}(x)
\end{equation}
where $\alpha^{k}$ is a stepsize chosen by one of the stepsize rules. The matrix $\nabla f_{y_{t}}(x) \nabla f_{y_{t}}(x)^{T}$ is a symmetric matrix certainly, so there is a matrix $\Delta^{k}=-\lambda_{\min }\left(\nabla f_{y_{t}}(x) \nabla f_{y_{t}}(x)^{T}\right) I$ that is a diagonal matrix that makes $\nabla f_{y_{t}}(x) \nabla f_{y_{t}}(x)^{T}+\Delta^{k}$ positive definite, as shown in Algorithm\ref{algorithm3}.

\begin{algorithm}[t] 
\caption{ Generate adversarial examples based on the Gauss-Newton Method} 
\label{algorithm3} 
\begin{algorithmic}[1] 
\REQUIRE $x$, $y_{t}$, $C$, $\delta  $, $l $

\ENSURE  $x$
\STATE  $\delta\leftarrow0$, $l\leftarrow y$
\STATE $\textbf {while}$\ $ l\neq y_{t}$ $\textbf {and}$   $\|\delta\|_{p} \leq C$  $\textbf {do}$
\label{code:fram:trainbase}

\STATE \ \ \ $\nabla f_{y_{t}}(x) \leftarrow\left[\frac{\partial f_{y_{t}}(x)}{\partial x_{i}}\right]_{\operatorname{m \times 1}}$\\
\STATE \ \ \  $H \leftarrow \nabla f_{y_{t}}(x) \cdot \nabla f_{y_{t}}(x)^{T}$\\
\STATE \ \ \  $\Delta^{k} \leftarrow-\lambda_{\min }\left(\nabla f_{y_{t}}(x) \nabla f_{y_{t}}(x)^{T}\right) I$
\STATE \ \ \  $\delta^{k+1}=\delta^{k}-\alpha^{k}\left(H+\Delta^{k}\right)^{-1} \nabla f_{y_{t}}(x) f_{y_{t}}(x)$

\STATE \ \ \ $x \leftarrow x+\delta$

\STATE \ \ \ $l \leftarrow Y\left(x\right)$

\STATE $\textbf {return}$\ \ $x $

\end{algorithmic}
\end{algorithm}

Although GN cannot produce closer adversarial examples since the substitution of the Hessian matrix $H+\Delta^{k}$ does not work as well as  the Hessian matrix, GN is better suited to generating synthetic digits that are meaningless to human eyes but still classified as each digit as shown in Fig 10. Sec.~\ref{Synthetic} performs generating synthetic digits using GN.

\section{EVALUATION}
\label{five}
We do some experimental evaluation to answer the following questions: (1) “In all possible objective functions, which one performs better?”, (2) “Compared with attacks that reported in previous literature, what's our advantage?”, (3) “Can we exploit any $x$?” and (4) “Can we defeat defensive distillation?”. Our main result is that TEAM can craft adversarial examples with a 100\% ASR by modifying the input on average by. Compared with these attacks, TEAM achieved highest ASR by 100\% with lowest distortion. We define a attack power measure to identify adversarial examples are more powerful than others.

\begin{figure}[t]
\begin{center}
\centerline{\includegraphics[width=8.5cm]{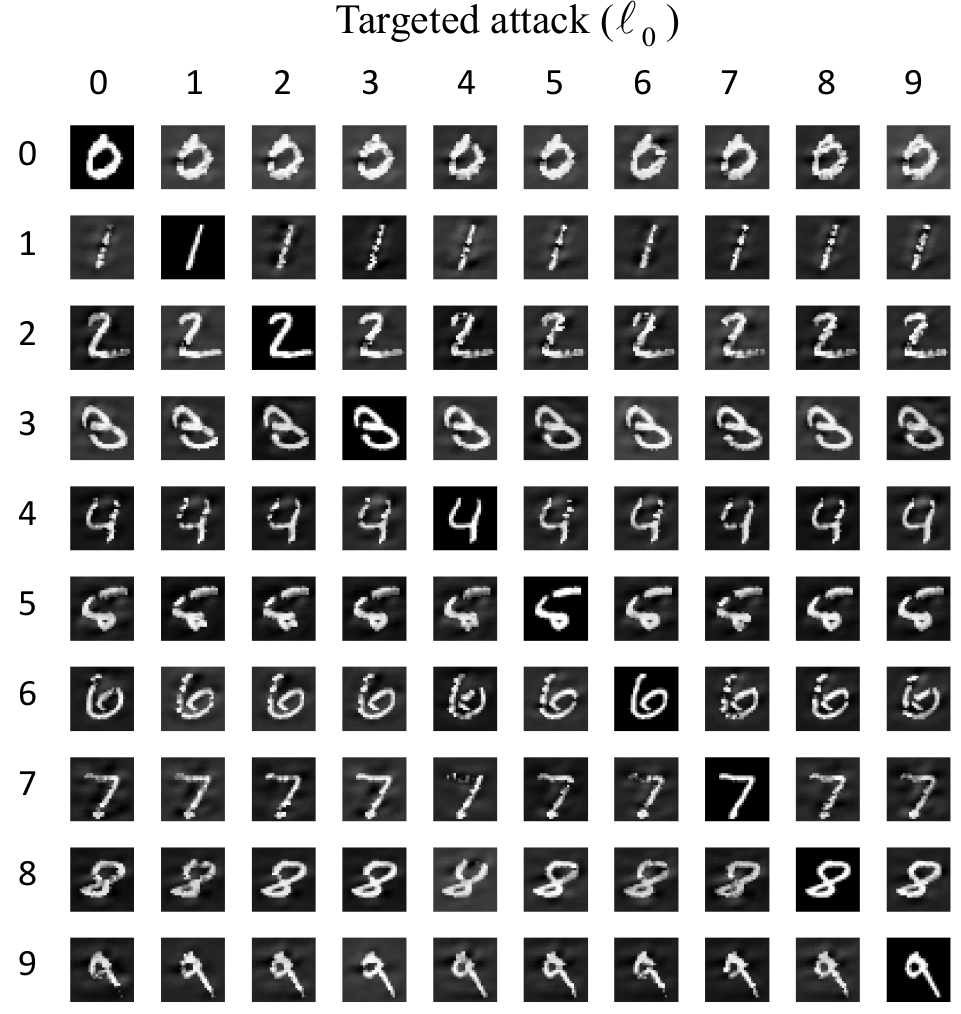}}
\caption{Our $\ell_{0}$ attack applied to the MNIST data set conducting a targeted attack for each target class.}
\label{mnsitl0}
\end{center}
\end{figure} 

We conducted our experiments on MNIST \cite{MNIST} and CIFAR10 \cite{CIFAR10} data sets. MNIST is widely used for digit-recognition tasks, which has a training set of 60,000 examples, and a test set of 10,000 examples. Each example includes a 28$\times$28 pixels grey image associated with its ground-truth label. The digits have been size-normalized and centered in a fixed-size image. CIFAR10 is suitable for a small image recognition task, which is divided into five training batches and one test batch, each with 10,000 examples. Each example in CIFAR10 consists of colour images of 3 channels with the size of 32$\times$32 pixels. We train two models, whose architecture are given in TABLE \ref{architecture}, on the MNIST and CIFAR10 and  achieve the classification accuracy of 99.3\% and 82\% respectively.

\begin{table}[H]

\begin{center}

\caption{Model architecture for  MNIST and CIFAR10 data sets}
\label{architecture}

\begin{tabular}{ccc}

\hline

 Layer Type           & MNIST Model & CIFAR10 Model  \\ \hline

Convolution + ReLU        & 3$\times$3$\times$32        &    3$\times$3$\times$64            \\ 
Max Pooling               & 2$\times$2          &         2$\times$2         \\ 
Convolution + ReLU        & 3$\times$3$\times$64        &       3$\times$3$\times$128            \\
Max Pooling               & 2$\times$2          &            2$\times$2           \\ 
Fully Connected + Sigmoid & 200         &        256       \\ 
Softmax                   & 10          &           10    \\ \hline
\end{tabular}
\end{center}
\end{table}

In addition, we use $\ell_{p}$, PSNR \cite{PSNR} and ASR to measure the effectiveness of TEAM. $\ell_{p}$ norm is a widely-used distance metric, often written as $\left\|x-x^{\prime}\right\|_{p}$, where $p$-norm $\left\| \cdot \right\|_{p}$ is defined as
\begin{equation}\|v\|_{p}=\left(\sum_{i=1}^{n}\left|v_{i}\right|^{p}\right)^{\frac{1}{p}}\end{equation}
More specifically, $\ell_{0}$ measures the number of $i$ such that $x_{i} \neq x_{i}^{\prime} $. $\ell_{2}$ measures the standard Euclidean distance between $x_{i}$ and $ x_{i}^{\prime}$. $\ell_{\infty}$ measures the maximum change of  $x_{i}$. Since no distance metric is a perfect measure of human perceptual similarity \cite{Carlini11}, we add PSNR as another distance metric.
We believe PSNR is an important indicator to measure $\delta$, since an image changed with larger PSNR value is less likely to be detected by human eyes. When ASR is not 100\%, $\ell_{p}$ and PSNR are used for successful attacks only. (That is, we only calculates  $\ell_{p}$ and PSNR value for adversarial examples that fooled DNN successfully.)

\begin{figure}[t]
\begin{center}
\centerline{\includegraphics[width=8.5cm]{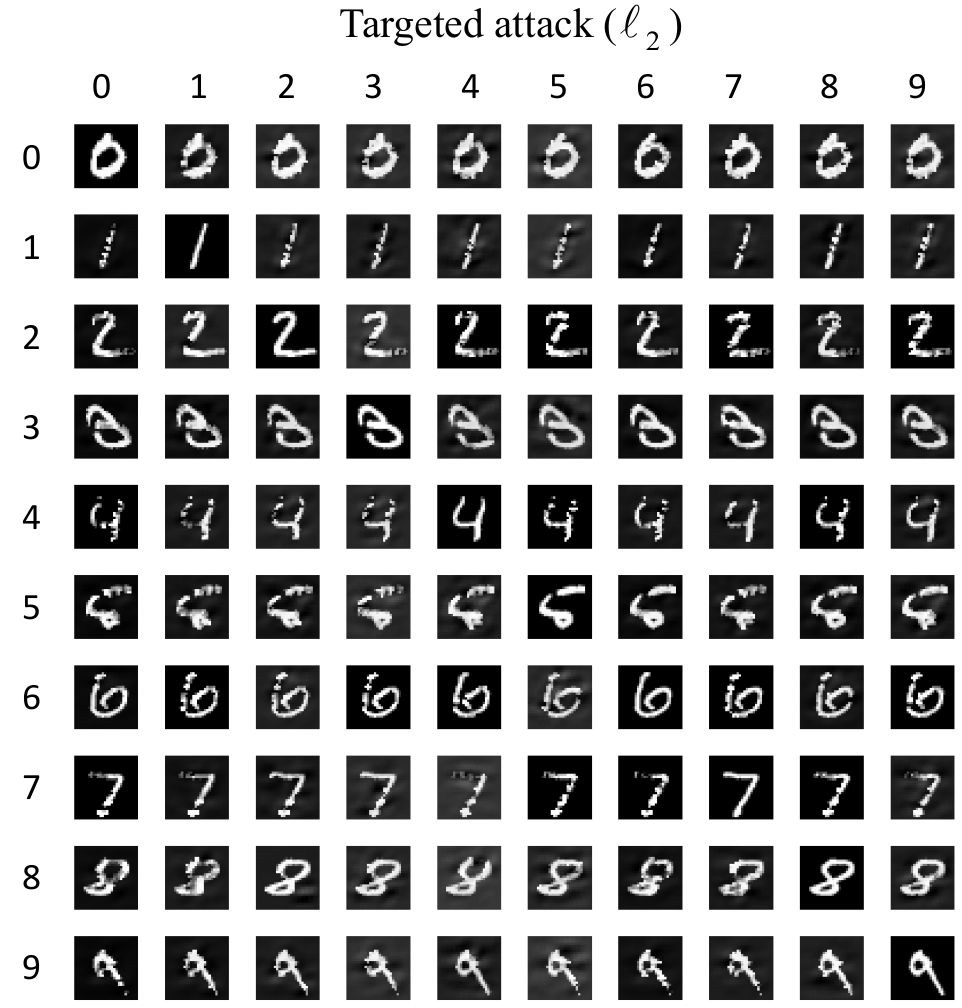}}
\caption{Our $\ell_{2}$ attacks applied to the MNIST data set conducting a targeted attack for each target class.}
\label{mnsitl2}
\end{center}
\end{figure} 
\subsection{Comparison Between Objective Functions}


\begin{table}[H]

\linespread{1.2}
\caption{Evaluation of untargeted attack by different objective functions on MNIST and CIFAR10. We show the average $\ell_{2}$ distortion, PSNR and ASR of the objective function $T_{1}$ and  $T_{4}$.}

\label{untargetedtable}
\begin{center}
\begin{small}
\begin{sc}
\begin{tabular}{ccccccc}
\hline

\multicolumn{1}{c}{} & \multicolumn{3}{c}{MNIST} & \multicolumn{3}{c}{CIFAR10} \\ 

                       & $\ell_{2}$     & \ \  PSNR    & ASR    & $\ell_{2}$      & PSNR     & ASR     \\ \hline

$T_{1}$                     & 0.91   &  \ \  73.03   & 100\%  & 0.29    & 87.41    & 100\%   \\

$T_{4}$                   & 1.40   &  \ \  71.46   & 100\%   & 0.30    & 87.45    & 100\%  \\ \hline

\end{tabular}
\end{sc}
\end{small}
\end{center}

\end{table}

\begin{table}[H]
\linespread{1.2}
\caption{Evaluation of targeted attack by different objective functions on MNIST and CIFAR10 data sets. We show the average $\ell_{2}$ distortion, PSNR and ASR of the objective function $T_{2}$, $T_{3}$, $T_{5}$ and  $T_{6}$.}
\label{targettable}
\begin{center}
\begin{small}
\begin{sc}
\scalebox{0.9}

\begin{tabular}{p{1.0cm}p{0.2cm}cp{0.6cm}ccp{0.6cm}c}
\hline

\multicolumn{2}{c}{\multirow{2}{*}{}}                                                                    & \multicolumn{3}{c}{MNIST} & \multicolumn{3}{c}{CIFAR10} \\
\multicolumn{2}{c}{}                                                                                     &  $\ell_{2}$     & PSNR   & ASR     & $\ell_{2}$     &   PSNR    & ASR      \\ 
\hline

\multirow{4}{*}{\begin{tabular}[c]{@{}c@{}}$Best$\\ $Case$\end{tabular}}  & $T_{2}$ & \textbf{1.12}   & \textbf{77.21}  & \textbf{100\%}   & \textbf{0.65}   & \textbf{74.83}   & \textbf{100\%}    \\
\multicolumn{1}{c}{}                                                                              & $T_{3}$ & 4.45   & 50.13  & 100\%   & 3.74   & 56.17   & 100\%    \\
\multicolumn{1}{c}{}                                                                              & $T_{5}$ & \textbf{1.23}   & \textbf{65.31}  & \textbf{100\%}   & \textbf{0.73}   & \textbf{70.12}   & \textbf{100\%}    \\ 
\multicolumn{1}{c}{}                                                                              & $T_{6}$ & 4.87   & 45.32  & 100\%   & 4.17   & 50.16   & 100\%    \\ \hline

\multirow{4}{*}{\begin{tabular}[c]{@{}c@{}}$Average$\\ $Case$\end{tabular}}                             & $T_{2}$ & \textbf{1.84}   & \textbf{67.21}  & \textbf{100\%}   & \textbf{0.70}   & \textbf{72.86}   & \textbf{100\%}    \\
                                                                                                    & $T_{3}$ & 5.12   & 45.26  & 75.2\%  & 4.52   & 46.71   & 70.1\%   \\
                                                                                                    & $T_{5}$ & \textbf{2.1}   & \textbf{69.15}  & \textbf{100\%}   & \textbf{1.3}   & \textbf{67.26}   & \textbf{100\%}   \\ 
                                                                                                    & $T_{6}$& 5.31   & 41.74  & 78.5\%  & 4.50   & 48.34   & 73.5\%   \\\hline

\multirow{4}{*}{\begin{tabular}[c]{@{}c@{}}$Worst$\\ $Case$\end{tabular}}                               &$T_{2}$ & \textbf{2.64}   & \textbf{58.75}  & \textbf{100\%}   & \textbf{0.75}   & \textbf{73.12}   & \textbf{100\%}    \\
                                                                                                    & $T_{3}$ & 6.54   & 39.17  & 74.5\%  &5.62   & 45.13   & 69.7\%   \\
                                                                                                    & $T_{5}$ & \textbf{3.41}   & \textbf{55.12}  & \textbf{100\%}   & \textbf{2.50}   & \textbf{60.73}   & \textbf{100\%}    \\ 
                                                                                                    & $T_{6}$ & 5.80   & 40.12  & 75.6\%  & 5.17   & 46.25   & 71.2\%  \\ \hline

\end{tabular}

\end{sc}
\end{small}
\end{center}

\end{table}

For objective function $T(\cdot)$, we evaluate the quality of the adversarial examples from each objective function on two data sets. $T_{1}$ and $T_{4}$ can both reach 100\%  ASR with lower distance metric and higher PSNR. Experimental results are shown in TABLE  \ref{untargetedtable} and TABLE  \ref{targettable}. TABLE  \ref{untargetedtable} is the results for untargeted attacks. The only difference between $T_{1}$ and $T_{4}$ is that the gradient of $T_{1}$ comes from the last hidden layer, while that of $T_{4}$ comes from the softmax layer. Since the experimental results obtained by $T_{1}$ and $T_{4}$ differed slightly, we conclude that unlike JSMA, extreme derivative values introduced by $S_{i}$ does not affect TEAM.

In TABLE  \ref{targettable}, we proposed three different approaches to choose the target class in targeted attacks: ``Average Case" means that we select the target class randomly among the set of labels that is not the ground-truth one. ``Best Case" means performing the attack against all the classes except for the ground-truth one and reporting the target class that is least difficult to attack. ``Worst Case" means performing the attack against all the classes and report the target class that is the most difficult to attack. As shown in  TABLE  \ref{targettable}, $T_{2}$ and $T_{5}$ perform better than others. Besides, the only difference between $T_{2}$ and $T_{5}$ is that the gradient of $T_{2}$ comes from the last hidden layer and the gradient of $T_{5}$ comes from the softmax layer.  $T_{2}$ and $T_{5}$ always achieve  100\% ASR, which outperform $T_{3}$ and $T_{6}$, since the increase of $T_{3}$ and $T_{6}$ does not closely related to the confidence of target class. Results from TABLE  \ref{targettable} also show that extreme derivative values introduced by $S_{i}$ does not affect TEAM and TEAM performs well both on the last hidden layer and softmax layer , which means our method could defeat the defense methods based on gradient masking, such as defensive distillation \cite{Papernot15}.

\begin{figure}[t]

\begin{center}
\centerline{\includegraphics[width=8.5cm]{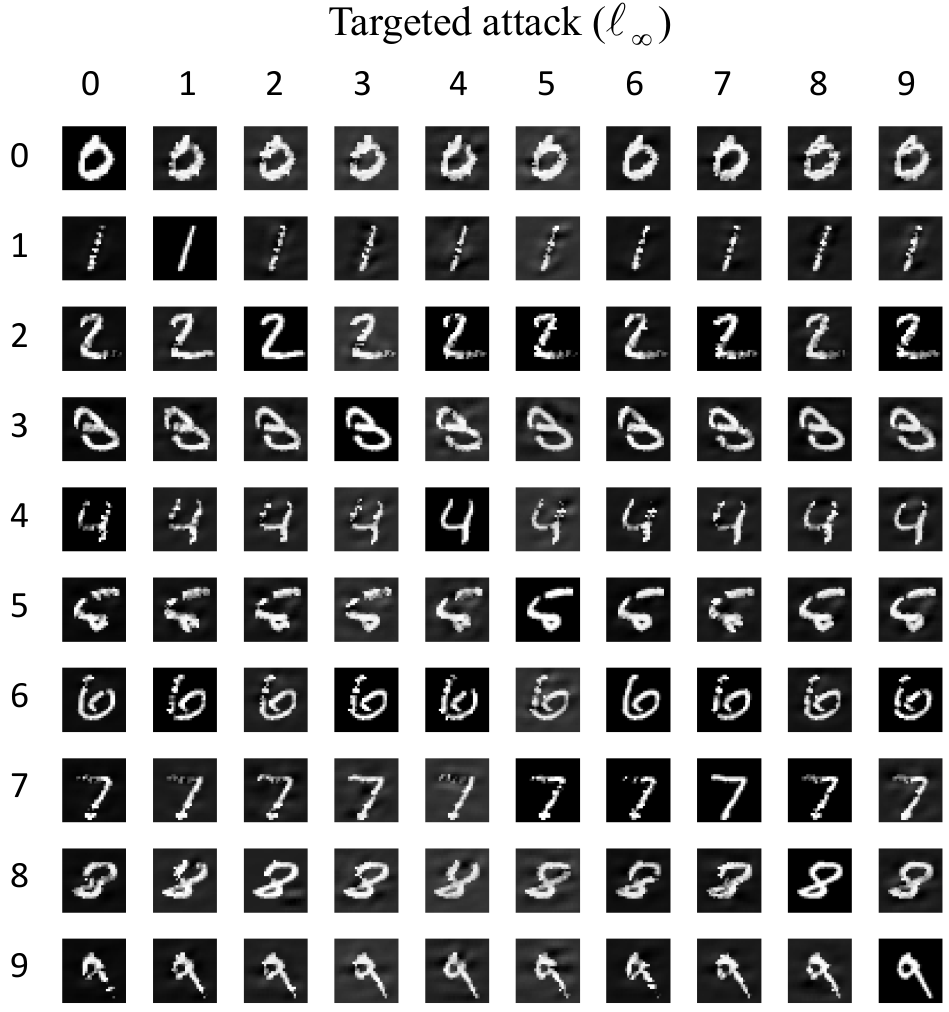}}
\caption{Our $\ell_{\infty}$ attack applied to the MNIST data set conducting a targeted attack for each target class.}
\label{mnsitlinf}
\end{center}
\vspace{-0.5cm}
\end{figure}

\subsection{Comparison Between Existing Classic Methods And TEAM}

To measure the effectiveness of TEAM, we compared the performance of TEAM with the previous classical targeted attacks, including JSMA($\ell_{0}$ norm), C\&W($\ell_{2}$ norm), FGSM($\ell_{\infty}$ norm) and untargeted attacks including Deepfool($\ell_{2}$ norm) and M-DI$^{2}$-FGSM($\ell_{2}$ norm)  on the same test data set.

We re-implement JSMA, C\&W, FGSM, Deepfool and M-DI$^{2}$-FGSM for comparison. The codes of JSMA, C\&W, FGSM and Deepfool come from Cleverhans \cite{Faghri26} and the codes for M-DI$^{2}$-FGSM come from the link given by the author in the original text \cite{Xie14}. In addition, to ensure the validity of the evaluation, we use the same model architecture in TABLE \ref{architecture} and the same batch of test data.  For JSMA, we make $\epsilon=1$ for every iteration and the stop condition is to meet a success attack. That is, under the targeted attack, we report metrics only if JSMA produces the target label, no matter how many perturbation pixels JSMA required. For C\&W,  we choose the $\ell_{2}$ attack and set the scaling factor $C=1.0$. For FGSM, we make perturbation bound $\varepsilon=0.01$, and epochs = 1. For Deepfool, we set epochs = 3. For M-DI$^{2}$-FGSM, we set the maximum perturbation bound of
each pixel $\varepsilon=0.05$ and decay factor is set to be 1.

For our experiments, we randomly selected  300 images  that could be correctly judged by the DNN in the test set on MNIST and CIFAR10. The results compared with targeted attacks are reported in TABLE \ref{targetedcompare} and the results compared with untargeted attacks are reported in TABLE \ref{untarge}. Fig.\ref{mnsitl0} to Fig.{cifarlinf} show our $\ell_{0}$ attacks, $\ell_{2}$ attacks and $\ell_{\infty}$ attacks applied to the two models for each target classes respectively. 

A powerful adversarial example should be an example with high ASR  and keep  less likely to be detected by human eyes(low $\ell_{p}$ value, high PSNR), and both two are indispensable. In TABLE \ref{targetedcompare}, for each distance metric, our attacks achieve 100\% ASR and keep  $\ell_{p}$ value lower than the state-of-the-art attacks with $\ell_{0}$ and  $\ell_{2}$ distance metric. Although prior  $\ell_{\infty}$ attacks find closer adversarial examples, our $\ell_{\infty}$ attack achieve higher ASR. In TABLE \ref{untarge}, our attacks find closer adversarial examples than the previous state-of-the-art attacks, and our attacks never fail to find an adversarial example.

\begin{figure}[t]
\begin{center}
\centerline{\includegraphics[width=8.5cm]{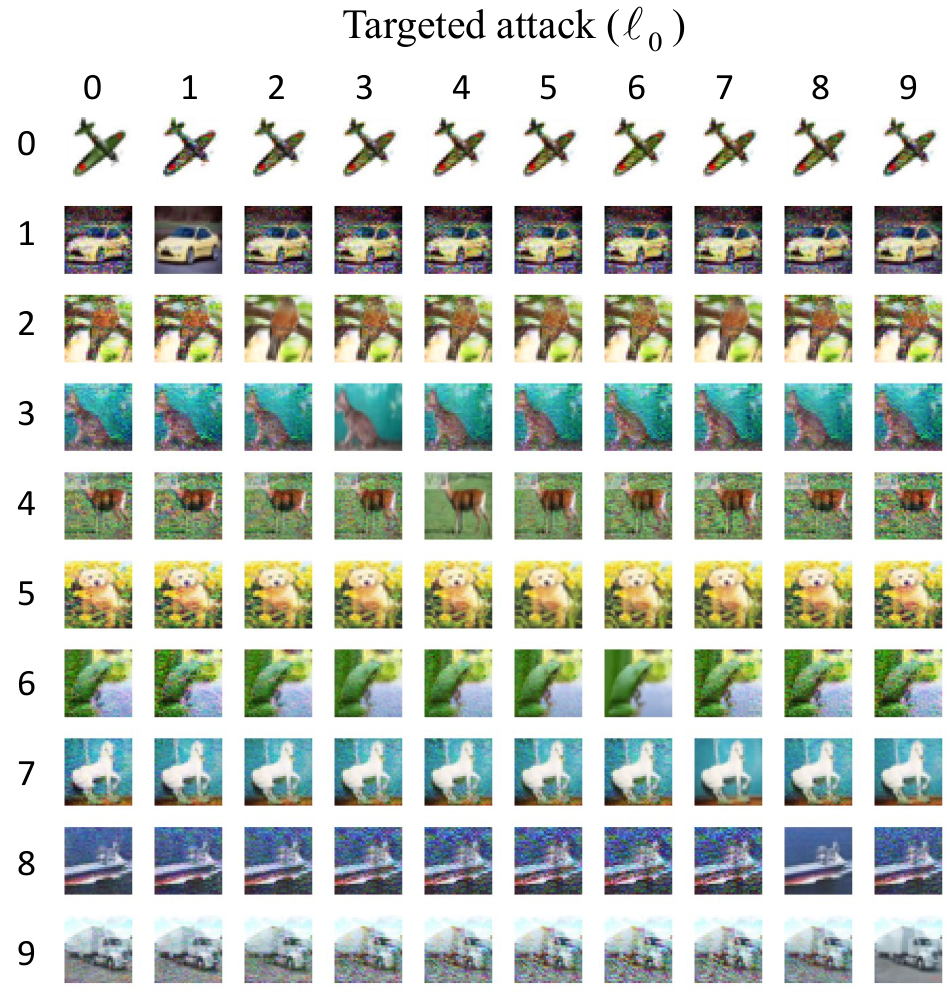}}
\caption{Our $\ell_{0}$ attack applied to the CIFAR10 data set conducting a targeted attack for each target class.}
\end{center}
\label{cifarl0}
\end{figure}

\begin{figure}[t]
\begin{center}
\centerline{\includegraphics[width=8.5cm]{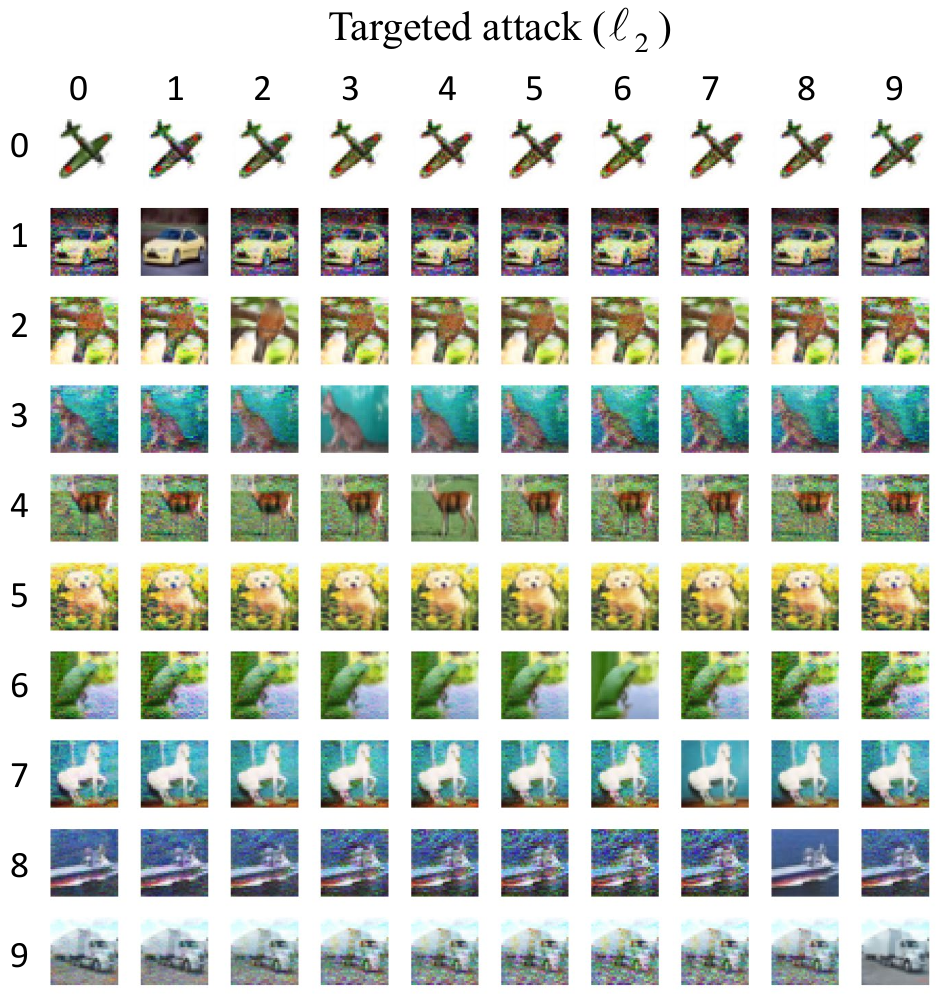}}
\vspace{-0.2cm}
\caption{Our $\ell_{2}$ attack applied to the CIFAR10 data set conducting a targeted attack for each target class.}
\end{center}
\label{cifarl2}
\end{figure}

\begin{table}[t]
\linespread{1.2}
\caption{Comparison of three targeted attacks algorithms on the MNIST and CIFAR10 data sets.}

\label{targetedcompare}
\vskip 0.05
in
\begin{center}
\begin{small}
\begin{sc}
\scalebox{0.9}

\begin{tabular}{ m{0.9cm}p{0.7cm}p{0.5cm}p{0.5cm}cp{0.5cm}p{0.5cm}c}
\hline

\multicolumn{2}{c}{\multirow{2}{*}{}}                                        & \multicolumn{3}{c}{MNIST} & \multicolumn{3}{c}{CIFAR10} \\
\multicolumn{2}{c}{}                                                         &  $\ell_{p}$     & PSNR   & ASR     &  $\ell_{p}$     & PSNR    & ASR      \\
\hline

\multirow{6}{*}{\begin{tabular}[c]{@{}c@{}}\\$Best$\\ $Case$\end{tabular}}     & $Our L_{0}$ & \textbf{20} & \textbf{71.21}  & \textbf{100\% }  & \textbf{21}      & \textbf{74.32}       &\textbf{100\%}        \\  
                                                                        & JSMA & 25  & 68.54  & 100\%   &25      & 70.16       & 100\%        \\ \cline{2-8} 
                                                                        &  $Our L_{2}$ & \textbf{1.12}   & \textbf{77.21}  &\textbf{100}\%   &\textbf{0.65}   & \textbf{74.83}   & \textbf{100}\%    \\ 
                                                                        & C\&W & 1.40   & 61.66  & 100\%   & 0.7   & 73.26   & 100\%    \\ \cline{2-8} 
                                                                        &  $Our \ell_{\infty}$ & \textbf{1.34}   & \textbf{70.86}  & \textbf{100\%}   &\textbf{1.32 }  & \textbf{70.56 }  &\textbf{ 100\%}    \\ 
                                                       & FGSM & 0.27   & 85.47  & 80\%  &0.036   & 91.72   & 85.3\%    \\
\hline

\multirow{6}{*}{\begin{tabular}[c]{@{}c@{}}\\$Average$\\ $Case$\end{tabular}} &$Our L_{0}$ & \textbf{56}  & \textbf{53.23} & \textbf{100\%}   & \textbf{93}      & \textbf{66.73}      & \textbf{100\%}        \\ 
                                                                        & JSMA & 76  & 51.49  & 100\%   &115      & 63.12       & 100\%        \\ \cline{2-8} 
                                                                        & $Our \ell_{2}$ & \textbf{1.84}  & \textbf{67.21}  & \textbf{100\%}  & \textbf{0.70}  & \textbf{72.86 }   & \textbf{100\%}    \\ 
                                                                        & C\&W & 2.21   & 60.68  & 96.2\%  & 0.75   & 68.52   & 99.2\%    \\ \cline{2-8}
                                                                        & $Our \ell_{\infty}$ & \textbf{2.32}   & \textbf{70.90}  & \textbf{100\%}   & \textbf{2.10}   & \textbf{69.74}   & \textbf{100\%}    \\

                                                                        & FGSM & 0.36   & 83.10  & 45.3\%  & 0.085   & 89   & 31.7\%   \\
\hline

\multirow{6}{*}{\begin{tabular}[c]{@{}c@{}}\\ $Worst$\\ $Case$\end{tabular}}   & $Our \ell_{0}$ & \textbf{180}  &\textbf{43.16}  & \textbf{100\%}   & \textbf{256}      & \textbf{53.19}       & \textbf{100\%}       \\
                                                                        &JSMA & 183  & 40.19  & 100\%   & 271     & 51.23       & 100\%        \\ \cline{2-8} 

                                                                        & $Our \ell_{2}$ & \textbf{2.64}   & \textbf{58.75}  & \textbf{100\%}   & \textbf{0.75}   & \textbf{66.13}   & \textbf{100\%}    \\
                                                                        & C\&W & 3.30   & 50.80  & 93.7\%   & 0.83   & 69.54   & 98.8\%    \\ \cline{2-8}
                                                                        &$Our \ell_{\infty}$&\textbf{3.10}    & \textbf{61.34}   & \textbf{100\%}    & \textbf{3.2}     & \textbf{65.12}    & \textbf{100\%}    \\

                                                                        & FGSM & -   & -  & 0\%  & 0.34   & 80.12  & 1\%  \\
\hline

\end{tabular}

\end{sc}
\end{small}
\end{center}

\end{table}

\begin{figure}[t]
\begin{center}
\centerline{\includegraphics[width=8.5cm]{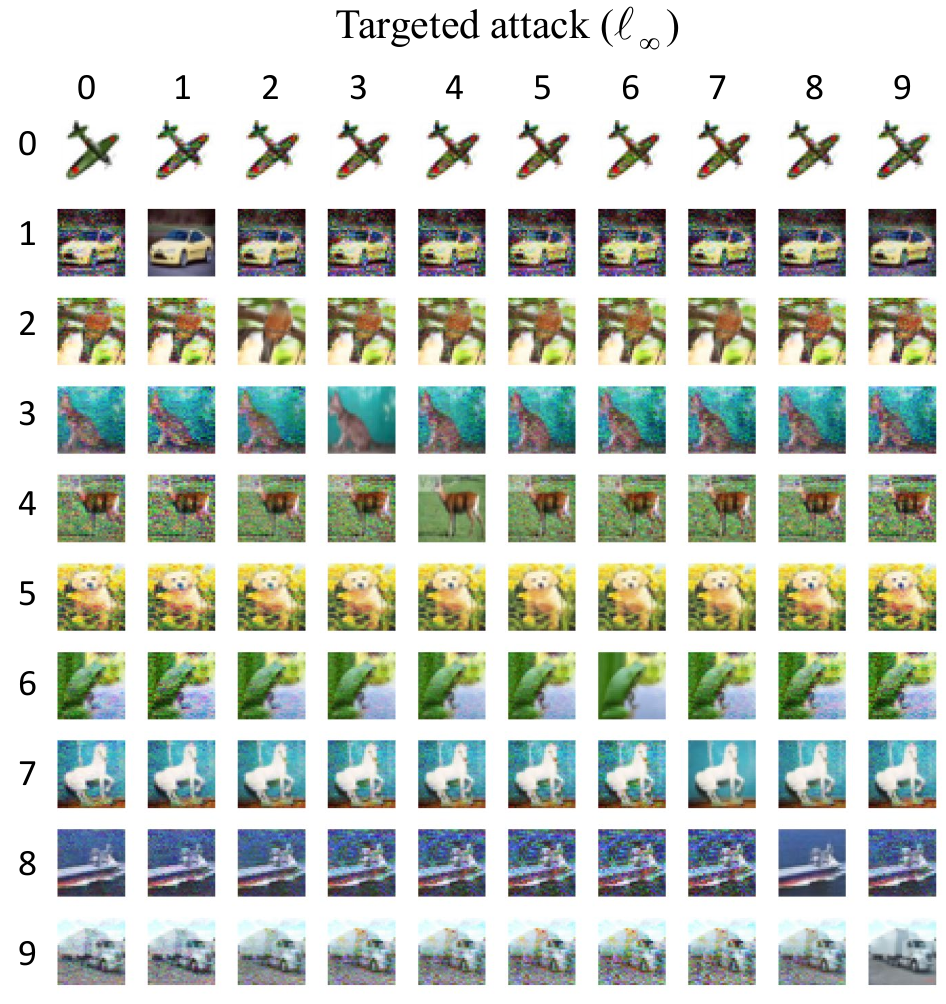}}
\caption{Our $\ell_{\infty}$ attack applied to the CIFAR10 data set conducting a targeted attack for each target class.}
\end{center}
\label{cifarlinf}
\end{figure}

\begin{table}[t]
\linespread{1.2}
\caption{Comparison between untargeted attack algorithms on the MNIST and CIFAR10 data sets.}

\label{untarge}
\vskip 0.05in
\begin{center}
\begin{small}
\begin{sc}
\scalebox{0.9}
{
\begin{tabular}{p{2.1cm}p{0.4cm}p{0.5cm}cp{0.4cm}p{0.5cm}c}
\hline

\multicolumn{1}{c}{} & \multicolumn{3}{c}{MNIST} & \multicolumn{3}{c}{CIFAR10} \\ 

                       & $\ell_{2}$     & PSNR    & ASR    & $\ell_{2}$      & PSNR     & ASR     \\ \hline
\ \ \ \ \ \ \ \ $T_{1}$                     & \textbf{0.91}   & \textbf{73.03}   & \textbf{100\%}  &  \textbf{0.29 }   & \textbf{87.41 }    & \textbf{100\%}   \\

\ \ \ \ \ \ \ \ $T_{4}$                     & \textbf{1.40}   & \textbf{71.46}   & \textbf{100\%}  & \textbf{0.30}    & \textbf{87.45}    & \textbf{100\%}   \\

\ \ \ \  ${\rm Deepfool}$                     & 2.21   & 73.80   & 100\%  & 0.9    &85.20    & 99.6\%   \\

\footnotesize M-DI$^{2}$-FGSM                   & 3.14   & 65.12   & 100\%   & 1.85   &75.12    & 99.4\%  \\ \hline

\end{tabular}
}
\end{sc}
\end{small}
\end{center}
\end{table}

JSMA uses the last hidden layer instead of the softmax layer to calculate the adversarial saliency map. Although the last hidden layer does not lead to extreme derivative values, it has its own properties that the increase of $Z_{t}$ does not bring the decrease of  $\sum_{i \neq t} Z_{i}$. Therefore,  the pixels selected by Eq.(\ref{jsma}) are not technically the key pixels in the targeted attack, which leads to larger $\ell_{0}$. Fortunately, our method is not affected by either derivative values of the softmax layer or properties of the last hidden layer, since Lagrangian multiplier method is able to achieve the optimal solution approximately to the origin problem regardless of $S$ and $Z$. 

C\&W and our method have similar objective functions, while our method use the second-order Taylor expansion to improve objective functions and use Lagrangian multiplier method instead of gradient descent  to optimize them. Compared to traditional hill climbing methods, the Lagrangian multiplier method can achieve better optimization results. FGSM is known as a weaken attack that aims at generating a large number of adversarial examples in a short time. It cares less about ASR, which leads to 0\% ASR in the worst case. 

Deepfool and M-DI$^{2}$-FGSM underestimate the fact that DNN are neither smooth nor convex, which leads to discontinuous and shattered gradients. And the discontinuous or shattered gradients likely lead to a local optimal solution. While our method use the second-order Taylor expansion to approximate the output value of the last hidden layer or softmax layer locally, which can overcome the discontinuous or shattered gradients problem.

\subsection{Generating Synthetic Digits}
\label{Synthetic}
\begin{figure}[h]

\begin{center}

\centerline{\includegraphics[width=9cm]{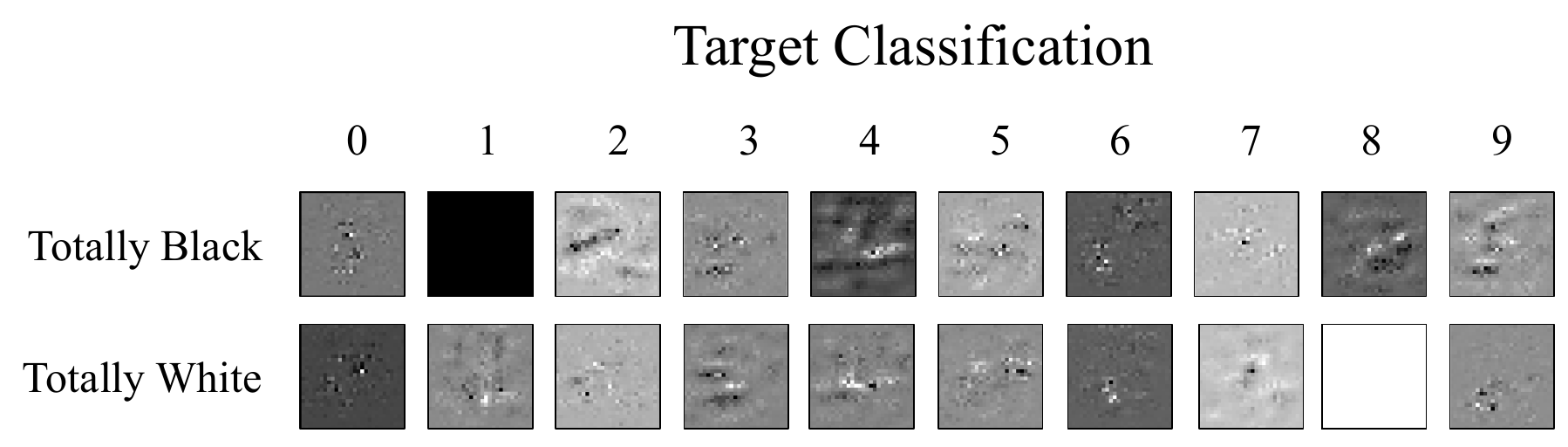}}
\caption{The targeted attack for the MNIST where the starting image is totally black or white.}

\label{GNpic}
\end{center}

\end{figure} 

With our targeted attack, we can begin with any input even it makes no sense to human beings and find adversarial examples of any target classes. Based on this, in Fig.\ref{GNpic} we show the results of using TEAM and GN to generate adversarial examples and make them classify as each class with respect to an entirely-black and entirely-white image. Papernotet al. \cite{Nicolas10} performed  $\ell_{0}$ task before, however, for classes 0, 2, 3 and 5, one can recognize the target class. While  none of the classes can be recognized with our more powerful attacks.
 
Notice that the entirely-black image requires no change to become class  and  the entirely-black image requires no change to become class  because they are initially classified as class 1 and class 8.

\subsection{Applying Our Attacks to Defensive distillation}

Hinton et al. \cite{Hinton31} proposed \textit{distillation} to transfer knowledge from a large model (teacher model) to a distilled model (student model). Papernot et al. \cite{Papernot15} exploited the notion of distillation to make the DNN robust against adversarial examples, namely \textit{defensive distillation}. They extract the knowledge in the form of class probability vectors of the training data from the teacher model and feed back to train the student model. To be more specific, they use the teacher model to label all the examples in the training set with their output vectors, namely soft label. Then, they use  the soft labels instead of the original hard labels to train the student model. Distillation increase the accuracy on the test set as well as the speed of  training the student model \cite{Hinton31,Melicher32}. In defensive distillation, the teacher model shares the same size with the student model and a distillation temperature constant $G$ is used to force the student model to become more confident in its predictions.

Defensive distillation modifies the softmax function by adding a temperature constant $G$:
\begin{equation}S_{i}(x, G)=\frac{\exp({z_{i} / G})}{ \sum_{j=1}^{m} \exp({z_{j} / G})}\end{equation}A larger $G$ forces the DNN to produce $S_{i}$ with relatively large values for each class. Since $Z_{i}$ become negligible compared to temperature $G$, $S_{i}$ converge to $1/N$ when $G \rightarrow \infty$. In other words, the larger the value of $G$ is, the more ambiguous $S_{i}$ will be, because all $S_{i}$ are close to  $1/N$. Defensive distillation proceeds in four steps: (1) train a teacher model with temperature $G$; (2) calculate soft labels from the teacher model; (3) train the student model with soft labels trained at temperature $G$; and (4) test the student model with  $G=1$.

\begin{figure}[t]

\centering

\subfigure[MNIST]{

\label{a}
\includegraphics[width=8cm]{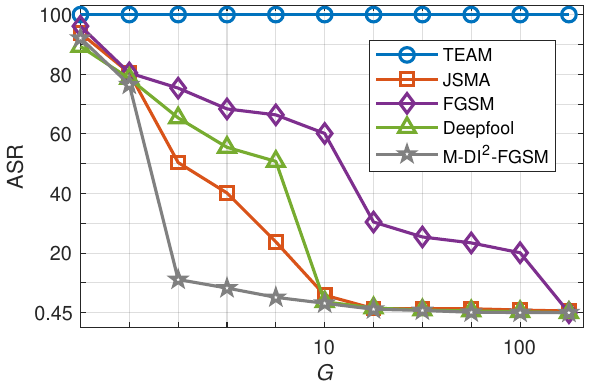}}

\hspace{1in}

\subfigure[CIFAR10]{
\label{b}

\includegraphics[width=8cm]{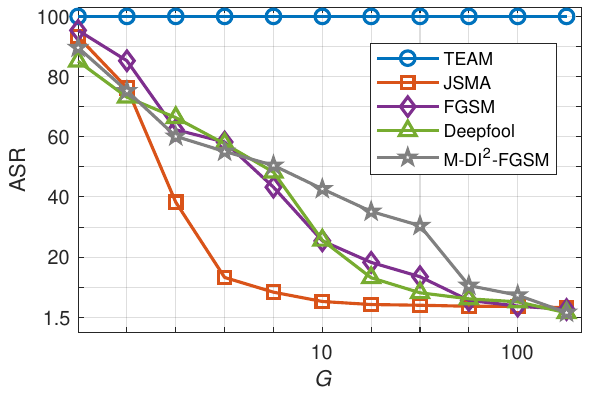}}

\caption{ASR of several typical adversarial attacks on a distillation defense with different distillation temperatures}

\label{distillationcompare}
\end{figure}

In \cite{Papernot15}, increasing $G$ was found to reduce ASR effectively. It goes from ASR = 95.89\% when $G=0$ to ASR $=0.45\% $ when $G=100$ on MNIST data set. We re-implement this experiment on MNIST and CIFAR10 with our attacks. We repeat the attack and vary the distillation temperature each time. The value of ASR for the following distillation temperatures $G$:\{1, 2, 4, 6, 8, 10, 20, 40, 60, 80, 100\} is measured. To compared with other attacks, we re-implement the same experiments with JSMA, FGSM, Deepfool and M-DI$^{2}$-FGSM. We didn't compare with C\&W, since C\&W has been proved in \cite{Carlini11}, which could achieve 100\% ASR. We train DNN with the temperature $G$ varied from $G=0$ to $G=100$. Fig.(\ref{distillationcompare})  plots the value of ASR with respect to temperature $G$ against different attacks where $\ell_{0}$ attacks keep $\|\delta\|_{0}\leq112$, $\ell_{2}$ attacks keep $\|\delta\|_{2}\leq 3.0$ and $\ell_{\infty}$ attacks keep $\|\delta\|_{\infty}\leq 0.3$.

Fig.(\ref{distillationcompare}) shows that defensive distillation turns out to be very effective to most of existing attacks, including JSMA, FGSM, Deepfool and M-DI$^{2}$-FGSM. However, defensive distillation provides only marginal value to our attacks. All of our attacks succeed with 100\% ASR.

Furthermore, we investigate the reason why distilled network make JSMA, FGSM, Deepfool and M-DI$^{2}$-FGSM attacking harder. When a distilled network was trained at temperature $G>1$ and test at temperature $G=1$, an increasing $G$ will reduce the absolute value of all components of $\nabla_{x}S$ \cite{Papernot33}. When $G=100$, FGSM, Deepfool and M-DI$^{2}$-FGSM are all failed due to the fact that  all components of $\nabla_{x}S$ are zero. However, JSMA fails for another reason. JSMA treats all $x_{i}$ equally, regardless of how much they change the softmax output \cite{Carlini11}. Therefore, JSMA fail to choose the  $x_{i}$ that leads to the most changes.

However, our objective functions doesn't rely on $\nabla_{x}S$, which make our attacks stay powerful to defense method based on gradient masking.

\section{CONCLUSION}
\label{six}

For evaluating the effectiveness of defense or training more robust DNNs, a powerful adversarial example is necessary. Different from previous first-order methods, we propose a novel second-order method namely TEAM, which approximates the output value of the DNN in a tiny neighborhood with regard to the  input. For generating synthetic digits, an improved version of TEAM using GN can generate adversarial  examples quickly without calculating the Hessian matrix. The experimental results show that TEAM achieves higher ASR with fewer perturbations compared with previous methods both in targeted attacks and untargeted attacks. Compare to the state-of-the-art method C\&W, our generated adversarial example keep  less  likely  to  be  detected  by  human eyes with a higher ASR. Besides, adversarial examples generated by TEAM defeated the defensive distillation that based on the gradient masking. In future work, we plan to propose more powerful attacks based on the second-order gradient with less computation to evaluate the robustness of DNNs directly.

\section*{Acknowledgment}

This work is supported by National Key R\&D Program (No. 2018YFB2100400), Zhejiang Provincial Natural Science Foundation of China (No. LY17F020011, No. LY18F020012), the Scientiﬁc Project of Zhejiang Provincial Science and Technology Department (No. LGG19F030001, No. LGF20F020007), Natural Science Foundation of China (No. 61902082, No. 61972357), and Zhejiang Key R\&D Program (No. 2019C03135).

\ifCLASSOPTIONcaptionsoff
  \newpage
\fi

\end{document}